%% file: main.tex
\documentclass[runningheads]{llncs}

\usepackage[mobile]{eccv}

\usepackage{eccvabbrv}

\usepackage{graphicx}
\usepackage{booktabs}
\usepackage{multirow}
\usepackage{amsmath}
\usepackage{amsfonts}
\usepackage{array} 
\usepackage{bbm}
\usepackage{threeparttable}
\usepackage{xcolor, colortbl}
\usepackage{pdfpages}
\usepackage{geometry}
\usepackage{lipsum}
 \usepackage{float}

\usepackage[accsupp]{axessibility}  %

\usepackage[pagebackref,breaklinks,colorlinks,citecolor=eccvblue]{hyperref}

\usepackage{orcidlink}

\input{shortcut}

\newcommand\blfootnote[1]{%
  \begingroup
  \renewcommand\thefootnote{}\footnote{#1}%
  \addtocounter{footnote}{-1}%
  \endgroup
}

\begin{document}

\title{GeoWizard: Unleashing the Diffusion Priors for 3D Geometry Estimation from a Single Image} 

\titlerunning{GeoWizard}

\author{Xiao Fu\inst{1*} \and
Wei Yin\inst{2*} \and 
Mu Hu\inst{3*} \and
Kaixuan Wang\inst{3} \and
Yuexin Ma\inst{4} \and \\
Ping Tan\inst{3,6} \and
Shaojie Shen\inst{3} \and
Dahua Lin\inst{1\text{†}} \and
Xiaoxiao Long\inst{5,6\text{†}}
}

\authorrunning{Fu. et al.}

\institute{
\vspace{-0.5em}\href{https://fuxiao0719.github.io/projects/geowizard/}{\textcolor[rgb]{0.5,0.5,0}{\textbf{https://fuxiao0719.github.io/projects/geowizard/}}} \\
\vspace{0.5em}$^{1}$CUHK $^{2}$The University of Adelaide $^{3}$HKUST \\ 
$^{4}$ShanghaiTech University $^{5}$HKU $^{6}$Light Illusions
}

\makeatletter
\let\@oldmaketitle\@maketitle%
\renewcommand{\@maketitle}{\@oldmaketitle%
 \centering
    \includegraphics[width=\linewidth]{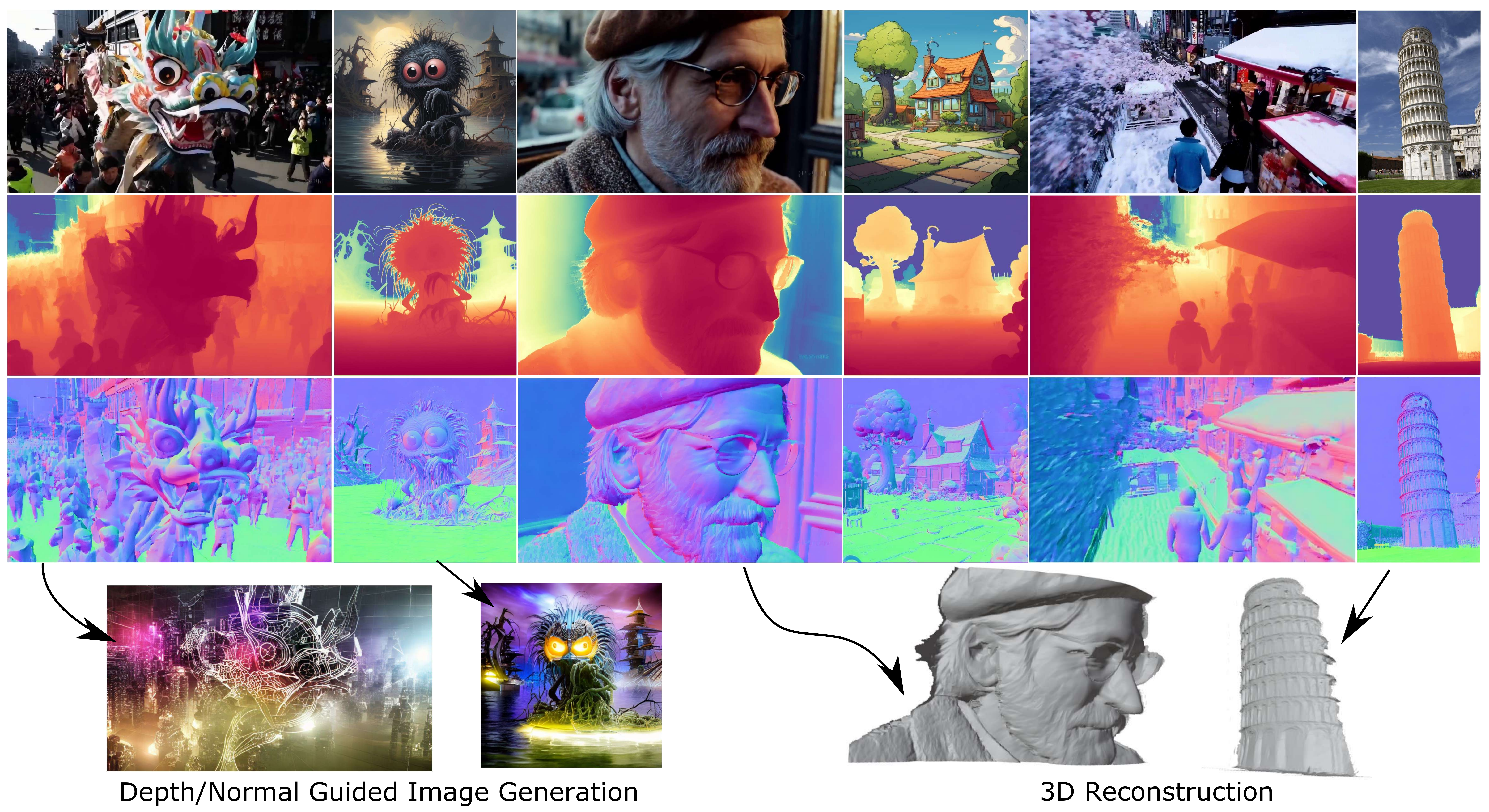}
     \captionof{figure}{We propose~\textit{GeoWizard}, an innovative foundation model for jointly estimating depth and surface normal from monocular images. Compared to prior discriminative counterparts, our work not only achieves surprisingly robust generalization on various types of real or unreal images but also faithfully captures intricate geometric details. The generated depth and normal could enhance many applications such as 2D content generation, 3D reconstruction and so on.
     \vspace{-2em}}
    \label{Fig: first page fig.}
    \bigskip} 
    
\makeatother

\maketitle

\input{main/0_abstract}
\input{main/1_intro}

\input{main/2_related}
\input{main/3_method}

\input{main/4_experiment}
\input{main/5_conclusion}
\input{main/6_acknowledgement}
\input{main/7_supp}

\clearpage
\bibliographystyle{splncs04}
\bibliography{main}

\end{document}

%% file: shortcut.tex
\newcommand{\bK}{\mathbf{K}}

\newcommand{\bk}{\mathbf{k}}
\newcommand{\bx}{\mathbf{x}}

\newcommand{\bs}{\mathbf{s}}

\newcommand{\bV}{\mathbf{V}}

\newcommand{\bn}{\mathbf{n}}
\newcommand{\bv}{\mathbf{v}}
\newcommand{\bq}{\mathbf{q}}

\newcommand{\bi}{\mathbf{i}}

\newcommand{\bz}{\mathbf{z}}

\newcommand{\bZ}{\mathbf{Z}}

\newcommand{\bQ}{\mathbf{Q}}

\newcommand{\bd}{\mathbf{d}}

\makeatletter
\DeclareRobustCommand\onedot{\futurelet\@let@token\@onedot}
\def\@onedot{\ifx\@let@token.\else.\null\fi\xspace}
\def\eg{e.g\onedot} 
\def\ie{i.e\onedot}

\def\etal{\textit{et~al}\onedot} 
\def\Fig{Fig\onedot}   
\makeatother

\newcommand{\figref}[1]{\Fig~\ref{#1}}
\newcommand{\secref}[1]{Section~\ref{#1}}
\newcommand{\appref}[1]{Appendix~\ref{#1}}

\renewcommand{\eqref}[1]{Eq.~\ref{#1}}
\newcommand{\tabref}[1]{Table~\ref{#1}}

\newcommand*\rot{\rotatebox{90}}

\newif\ifcomment
\commenttrue
\ifcomment
	\newcommand{\yl}[1]{ \noindent {\color{cyan} {\bf Yiyi:} {#1}} }

\else
	\newcommand{\ag}[1]{}
	\newcommand{\yl}[1]{}
\fi

%% file: main/0_abstract.tex
\begin{abstract}
\blfootnote{$^{*}$ Equal contribution \quad $^{\text{†}}$ Corresponding author}We introduce \textit{GeoWizard}, a new generative foundation model designed for estimating geometric attributes,~\eg, depth and normals, from single images. While significant research has already been conducted in this area, the progress has been substantially limited by the low diversity and poor quality of publicly available datasets. As a result, the prior works either are constrained to limited scenarios or suffer from the inability to capture geometric details. In this paper, we demonstrate that generative models, as opposed to traditional discriminative models (\eg, CNNs and Transformers), can effectively address the inherently ill-posed problem. We further show that leveraging diffusion priors can markedly improve generalization, detail preservation, and efficiency in resource usage. Specifically, we extend the original stable diffusion model to jointly predict depth and normal, allowing mutual information exchange and high consistency between the two representations. More importantly, we propose a simple yet effective strategy to segregate the complex data distribution of various scenes into distinct sub-distributions. This strategy enables our model to recognize different scene layouts, capturing 3D geometry with remarkable fidelity. \textit{GeoWizard} sets new benchmarks for zero-shot depth and normal prediction, significantly enhancing many downstream applications such as 3D reconstruction, 2D content creation, and novel viewpoint synthesis.

\keywords{Monocular Images \and Depth \and Normal \and Diffusion Models}
\end{abstract}

%% file: main/1_intro.tex
\section{Introduction}

Estimating 3D geometry,~\eg, depth and surface normal from monocular color images, is a fundamental but challenging problem in 3D computer vision, which plays essential roles in various downstream applications such as autonomous driving~\cite{godard2017unsupervised,godard2019digging}, 3D surface reconstruction~\cite{yu2022monosdf,wang2022neuris,long2023wonder3d}, novel view synthesis~\cite{liu2022neural,pumarola2021d}, inverse rendering~\cite{yu2019inverserendernet,sengupta2019neural}, and so on.
Reverting the projection from a 3D environment to a 2D image presents a geometrically ambiguous challenge, necessitating the aid of prior knowledge. This may include understanding typical object dimensions and shapes, probable scene arrangements, as well as occlusion patterns.

The recent advancements in deep learning have significantly propelled the field of geometry estimation forward. Currently, this task is often approached as a neural image-to-image translation problem, where supervised learning techniques are employed. 
However, the progress in this area is constrained by two major shortcomings in the publicly available datasets:
1) \textbf{Low diversity.}
Lacking efficient and reliable tools for data collection, most datasets are confined to specific scenarios, such as autonomous driving and indoor environments. Models trained on these datasets typically exhibit poor generalization capabilities when applied to out-of-domain images.
2) \textbf{Poor accuracy.}
To enhance dataset diversity, some works generate pseudo labels for unlabeled data using methods like multi-view stereo (MVS) reconstruction or self-training techniques. Unfortunately, these pseudo-labels often suffer from being incomplete or of low quality. Consequently, while these approaches may improve model generalization, they still struggle in accurately capturing geometric details and require significantly more computational resources.

In this paper, our goal is to build a foundation model for monocular geometry estimation capable of producing high-quality depth and normal information for any images of any scenarios (even images generated by AIGC). 
Instead of employing straightforward data and computation scaling-up, our method proposes to unleash the diffusion priors for this ill-posed problem.
The intuition is that stable diffusion models have been proven to inherently encode rich knowledge of the 3D world, and its strong diffusion priors pre-trained on billions of images could significantly facilitate potential 3D tasks.

Instead of tackling depth or normal estimation separately, \textit{GeoWizard} jointly estimates depth and normal within a unified framework. Inspired by Wonder3D~\cite{long2023wonder3d}, we leverage \textbf{\textit{geometry switcher}} to extend a single stable diffusion model to produce both depth and normal. The joint estimation allows mutual information exchange and high consistency between the two representations.
However, direct training on mixed data encompassing various scenarios often leads to ambiguities in geometry estimation, potentially skewing the estimated depth/normal towards unintended layouts. To address this challenge, we propose a simple yet effective strategy, \textbf{\textit{scene distribution decoupler}}, to segregate the complex data distribution of different scenes into distinct sub-distributions (e.g., outdoor, indoor, and background-free objects). This strategic approach enables the diffusion model to discern different scene layouts, resulting in the capture of 3D geometry with remarkable fidelity.
Consequently, \textit{GeoWizard} achieves state-of-the-art performance in zero-shot depth and normal prediction, thereby significantly enhancing numerous downstream applications such as 3D reconstruction, 2D content creation, and novel viewpoint synthesis.

Overall, our contributions are summarized as follows:
\begin{itemize}
    \item We present {\textit{GeoWizard}}, a new generative foundation model for joint depth and normal estimation that faithfully captures intricate geometric details.
    \item We propose a simple yet effective \textit{scene distribution decoupler} strategy, aimed at guiding diffusion models to circumvent ambiguities that may otherwise lead to the conflation of distinct scene layouts.
    \item {\textit{GeoWizard}} achieves SOTA performance in zero-shot estimation of both depth and normal, substantially enhancing a wide range of applications.
\end{itemize}

%% file: main/2_related.tex
\section{Related Work}

\noindent \textbf{Joint Depth and Normal Estimation.} Estimating depth and normal from images is an ill-posed but important task, where depth and surface normal encode the 3D geometry in different aspects. 
Some existing approaches propose to explicitly acquire the surface normal from the depth map by using some geometric constraints, such as Sobel-like operator~\cite{hu2019revisiting,kusupati2020normal}, differentiable least square~\cite{long2020occlusion,qi2018geonet}, or randomly sampled point triplets~\cite{yin2021virtual,yin2019enforcing, long2024adaptive}. IronDepth~\cite{bae2022irondepth} propagates depth on pre-computed local surface. Zhao~\etal~\cite{zhao2021confidence} proposes to jointly refine depth and normal by a solver, but it conditions on multi-view prior and tedious post-optimization. On the other hand, %
several works~\cite{eigen2015predicting,li2015depth,xu2018pad,zhang2019pattern} create multiple branches for depth and normal, and enforce information exchange through propagating latent features. 
However, all the prior works tackle this problem using discriminative models and leverage limited scopes of training datasets, and therefore present poor generalization and fail to capture geometric details.
In contrast, \textit{GeoWizard} builds on generative models and fully leverage diffusion priors to tackle this problem, showing significantly improved generalization and ability to capture geometric details.

\subsubsection{\textbf{Diffusion Models for Geometry Estimation.}} Recently, diffusion models~\cite{ho2020denoising,sohl2015deep} have shown supreme capabilities in 2D image generation~\cite{dhariwal2021diffusion,rombach2022high,zhang2023adding,peebles2023scalable}. In constrast to GAN~\cite{bhattad2024stylegan}, some new works show that diffusion models can be employed in some 3D tasks, such as 
optical flow estimation~\cite{saxena2024surprising,dong2023openddvm}, view synthesis~\cite{liu2023zero,shi2023mvdream,sargent2023zeronvs}, depth estimation~\cite{ji2023ddp,ke2023repurposing,zhao2023unleashing}, and normal estimation~\cite{qiu2023richdreamer,long2023wonder3d,liu2023hyperhuman}.

For depth estimation, 
DDP~\cite{ji2023ddp} first introduces a unified diffusion architecture that blends the traditional perception pipeline to estimate the metric depth. %
DDVM~\cite{saxena2024surprising} further boosts depth quality by %
training on synthetic data. Although %
they leverage improved diffusion process~\cite{song2020denoising} or advanced perception backbone~\cite{liu2021swin,lin2017feature} to speed up training, they still suffer from unaffordable low efficiency and slow convergence when scaled up to internet-scale data. This is because these methods attempt to train diffusion models from scratch and ignore the strong diffusion priors of the pre-trained large diffusion models.
A concurrent method Marigold~\cite{ke2023repurposing} fine-tune the pre-trained stable diffusion model for depth estimation and also try to leverage the diffusion priors. However, it suffers from the ambiguities about mixed layouts of various scenarios, and tends to produce depth maps with unintended layouts.

Diffusion-based methods are also applied to normal estimation. JointNet~\cite{zhang2023jointnet} attempts to connect multiple diffusion models to achieve multi-modality estimation (e.g., depth and normal), however their model size and resource costs will linearly increase depending on the number of modalities. 
Wonder3D~\cite{long2023wonder3d} proposes to model joint color and normal distribution with a domain switcher to enhance geometric quality and consistency. 
Richdreamer~\cite{qiu2023richdreamer} trains seperately depth and normal diffusion model on the LAION-2B~\cite{schuhmann2022laion} dataset with predictions from Midas~\cite{Ranftl2022}.
However, these methods still struggle to capture geometric details. In contrast, to the best of our knowledge, \textit{GeoWizard} reveals robust generalization and significant ability to capture intricate geometric details.

%% file: main/3_method.tex
\section{Methodology}

Given an input image $\bx$, our goal is to generate 
its paired depth map $\hat{\bd}$ and normal map $\hat{\bn}$. 
Firstly, we delve into the problem with the diffusion paradigm (see \secref{sec:preliminaries}). Secondly, we present our geometric diffusion model (see \secref{sec:model}). The model uses a cross-domain geometry switcher to jointly generate the depth and normal using a single diffusion model. The mutual information exchange enhances geometric consistency. We further %
decouple the sophisticated scene distribution into several distinct sub-distributions (e.g., outdoor, indoor, and background-free objects) to avoid ambiguities of geometry estimation. 
Finally, the paired depth and normal are used for single-image-based 3D reconstruction (see \secref{sec:3drecon}). 
The overview of \textit{GeoWizard} is presented in \figref{fig:pipeline}.

\subsection{Preliminaries on Geometric Distribution} \label{sec:preliminaries}

Diffusion Probabilistic Models~\cite{ho2020denoising,sohl2015deep} define a forward Markov chain that progressively transits the sample $\bx$ drawn from data distribution $p ({\bx})$ into noisy versions $\{ \bx_{t}, t \in (1, T) | \bx_{t} = \alpha_{t} \bx_{0} + \sigma_{t} \boldsymbol{\epsilon} \}$, where $\boldsymbol{\epsilon} \sim \mathcal{N}(\mathbf{0}, \mathbf{I})$, $T$ is the training step, $\alpha_{t}$ and $\sigma_{t}$ are the noisy scheduler terms that control sample quality. In the reverse Markov chain, it learns a denoising network $\boldsymbol{\hat{\epsilon}}_{\boldsymbol{\theta}}(\cdot)$ parameterized by $\boldsymbol{\theta}$ usually structured as U-Net~\cite{ronneberger2015u} to transform $\bx_{t}$ into $\bx_{t-1}$ from an initial Gaussian sample $\bx_{T}$ through iterative denoising.

Unlike prior works that adopt CNN or transformer as architecture, we employ a diffusion-based scheme $f(\cdot)$ to model the joint depth and normal distribution $p({\bd}, {\bn})$. A 3D asset $\bZ$ possesses various attributes, such as albedo, roughness, and metalness, to describe its characteristics. We focus on depth and normal to represent the 3D spatial structure, approximating it to the distribution of a 3D asset $ p_{z} \approx p({\bd}, {\bn})$. Given a conditional input image $\bx$, the depth map $\hat{\bd}$ and the normal map $\hat{\bn}$ can be obtained by the generative formulation $f(\cdot) : \bx\in\mathbb{R}^3 \rightarrow (\hat{\bd}\in\mathbb{R}^+, \hat{\bn}\in\mathbb{R}^3)$, or in Markov probabilistic form:

\begin{equation}
\vspace{-5mm}
f(\bx) = p\left(\hat{\bd}_T, \hat{\bn}_T\right) \prod\limits_{t=1}^{T} p_{\boldsymbol{\theta}}\left(\hat{\bd}_{t-1}, \hat{\bn}_{t-1} \mid \hat{\bd}_{t}, \hat{\bn}_{t}, \bx \right)
\end{equation}
where $\hat{\bd}_T, \hat{\bn}_T \sim \mathcal{N}(\mathbf{0}, \mathbf{I})$. 

\begin{figure*}[tp!]
\centering
\includegraphics[width=\linewidth]{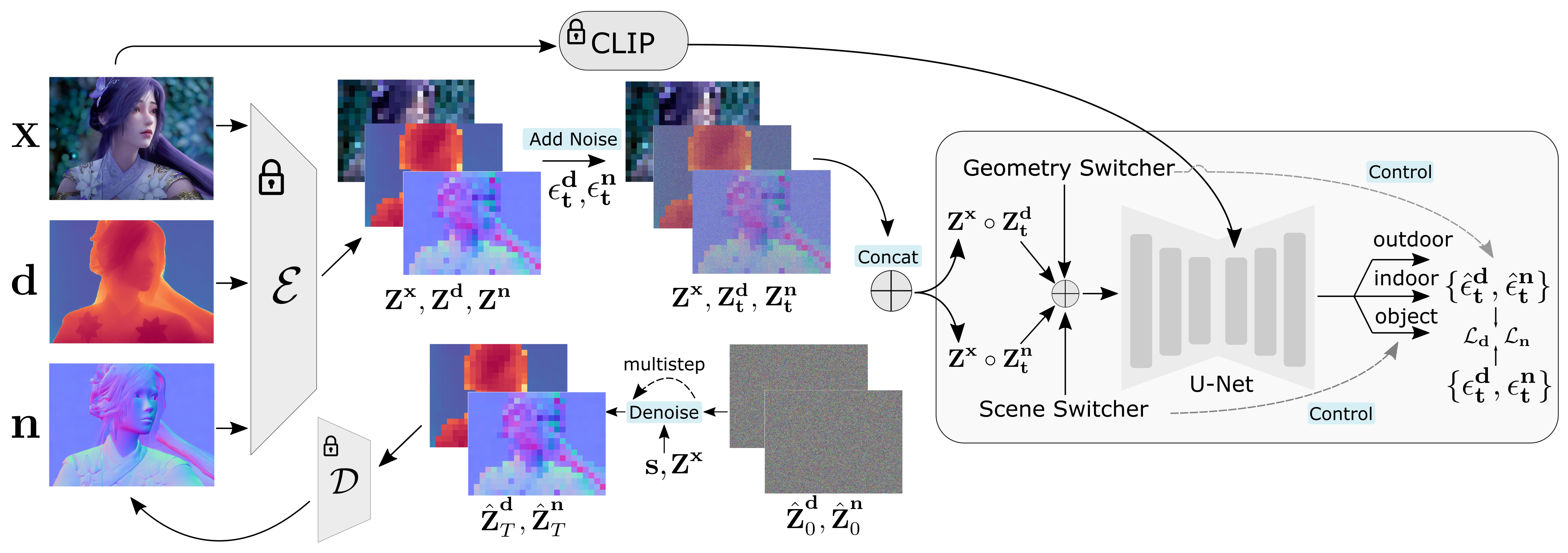}

\caption{\textbf{The overall framework of GeoWizard.} 
During fine-tuning, it first encodes the image $\bx$, GT depth $\bd$, and GT normal $\bn$ through the original stable diffusion VAE $\boldsymbol{\varepsilon}$ into latent space, yielding latents $\bZ^{\bx}$, $\bZ^{\bd}$, and $\bZ^{\bn}$ respectively. 
The two geometric latents are concatenated with $\bZ^{\bx}$ to form two groups, $\bZ^{\bx} \circ \bZ^{\bd}_{t}$ and $\bZ^{\bx} \circ \bZ^{\bn}_{t}$. 
Each group is fed into the U-Net to generate the output in depth or normal domain in the guide of a geometry switcher. 
Additionally, the scene prompt $\textbf{s}$ is introduced to produce results with one of three possible scene layouts (indoor/outdoor/object). During inference, given an image $\bx$, a scene prompt $\textbf{s}$, initial depth noise  $\boldsymbol{\epsilon}^{\bd}_{t}$ and normal noise  $\boldsymbol{\epsilon}^{\bn}_{t}$, GeoWizard can generate high-quality depth $\hat{\bd}$ and normal $\hat{\bn}$ jointly.}
\vspace{-2 em}
\label{fig:pipeline}
\end{figure*}

As shown in \figref{fig:pipeline}, the condition $\bx$ is integrated into the network in two ways: one is through the image embedding from CLIP~\cite{radford2021learning} for classifier-free guidance~\cite{ho2022classifier} via cross-attention layers, and the other is by concatenating it in the latent space with geometric latents for more precise control. Our intuition is that the CLIP embeddings offer global-wise guidance, enhancing the model robustness and expressiveness under various Gaussian initialization, while the latent-wise concatenation further reduces randomness when generating $\boldsymbol{\hat{\epsilon}}^{\bd}_{t}$ and $\boldsymbol{\hat{\epsilon}}^{\bn}_{t}$. Our main challenge is to characterize the distribution $p_{\boldsymbol{\theta}}$ or specifically $\boldsymbol{\hat{\epsilon}}_{\boldsymbol{\theta}}$ to generate high-quality depth and normal maps. 

\subsection{Geometric Diffusion Model} \label{sec:model}

Therefore, we base our model on the pre-trained 2D latent diffusion model (Stable Diffusion~\cite{rombach2022high}) so as to 1) utilize the strong, generalizable image priors learned from LAION-5B~\cite{schuhmann2022laion} 2) efficiently learn geometric priors in a low-dimensional latent space with minimum adjustments needed for U-Net architecture. 
However, this problem is non-trivial with two potential challenges: 1) the naive LDM is trained in the RGB domain, and thus may lack the capability to capture structural information and even impede it with reverse resistance. 2) The structure distributions are typically uniform, featuring similar values in localized areas, making them challenging for diffusion models to learn~\cite{lin2024common}.

\begin{figure*}[tp!]
\centering
\includegraphics[width=\linewidth]{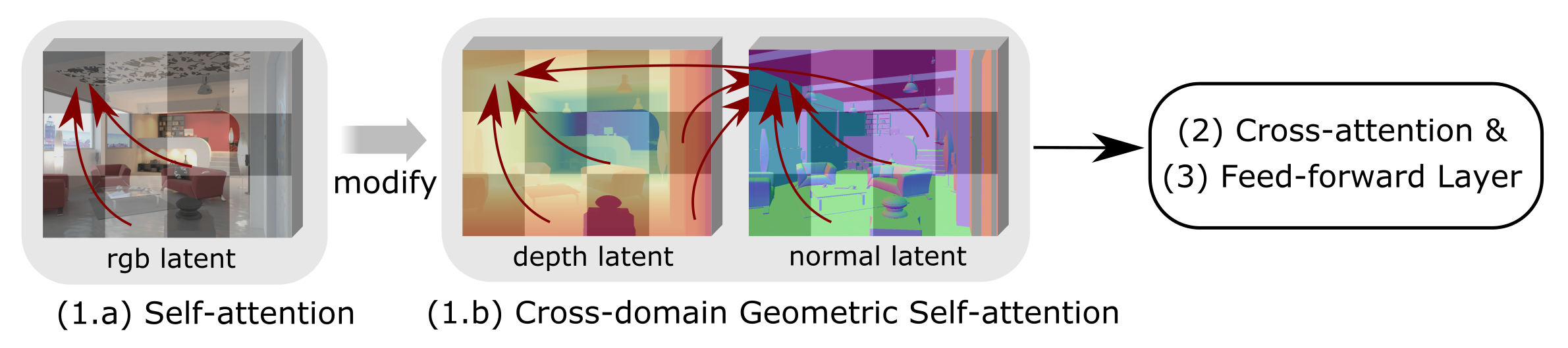}

\caption{\textbf{The Structure of Geometric Transformer Block.} Differing from the traditional self-attention layer (1.a) applied to RGB latent, we adapt it to a cross-domain geometric self-attention (1.b) that operates on depth latent and normal latent. This modification allows for mutual guidance and ensures geometric consistency.}
\vspace{-2 em}
\label{fig:crossdomain}
\end{figure*}

\noindent
\textbf{Joint Depth and Normal Estimation.}
To incorporate depth and normal for geometry estimation, one naive solution is to finetune two U-Nets ($f_{\bd}$, $f_{\bn}$) to model depth and normal distributions separately, \ie, $\hat{\bd}=f_{\bd}(\bx)$, $\hat{\bn}=f_{\bn}(\bx)$. However, this approach introduces extra parameters and overlooks the inherent connections between depth and normal, as both contribute to the unified geometric representation of a 3D shape. Normal describes surface variations and undulations %
, while depth outlines the spatial arrangement, guiding the orientation of normal. Our empirical experiment finds that this naive solution leads to geometric inconsistency in both depth and normal domain.

Inspired by~\cite{long2023wonder3d}, we leverage a geometry switcher to enable a single stable diffusion model to generate depth or normal through indicators. Specifically, $\hat{\bd}=f(\bx, \bs_{\bd})$, $\hat{\bn}=f(\bx, \bs_{\bn})$, where $\bs_{\bd}$ and $\bs_{\bn}$ are one-dimensional vectors that control depth and normal domain, respectively. The switchers are encoded by the low-dimensional positional encoding and added with time embedding in the U-Net. We find that using switchers converges faster than shared modeling~\cite{liu2023hyperhuman} or sequential modeling~\cite{long2023wonder3d}, and leads to more stable results.

To further enable mutual-guided geometric optimization, we modify the self-attention layer in U-Net to a cross-domain geometric self-attention layer to encourage spatial alignment, as shown in \figref{fig:crossdomain}. This operator not only improves geometric consistency between depth and normal but also leads to faster convergence. We compute queries, keys, and values as follows:
\begin{equation}
	\begin{split}
	& \bq_{\bd}=\bQ \cdot \hat{\bz}^{\bd}, \bk_{\bd}=\bK \cdot (\hat{\bz}^{\bd} \oplus \hat{\bz}^{\bn}), \bv_{\bd}=\bV \cdot (\hat{\bz}^{\bd} \oplus \hat{\bz}^{\bn})\\
	& \bq_{\bn}=\bQ \cdot \hat{\bz}^{\bn}, \bk_{\bn}=\bK \cdot (\hat{\bz}^{\bn} \oplus \hat{\bz}^{\bd}), \bv_{\bn}=\bV \cdot (\hat{\bz}^{\bn} \oplus \hat{\bz}^{\bd})
	\end{split}
\end{equation}
where $\hat{\bz}^{\bd}$ and $\hat{\bz}^{\bn}$ are latent depth and normal embeddings in transformer blocks, $\oplus$ denotes concatenation, and $\bQ$, $\bK$ and $\bV$ are query, key and value embeddings matrices. The cross-domain features are $\textbf{Att}(\bq_{i}, \bk_{i}, \bv_{i}), i = \{ \bd, \bn \}$, where $\textbf{Att}(\cdot)$ denotes softmax attention.

\begin{figure*}[tp]
\centering
\includegraphics[width=.95\linewidth]{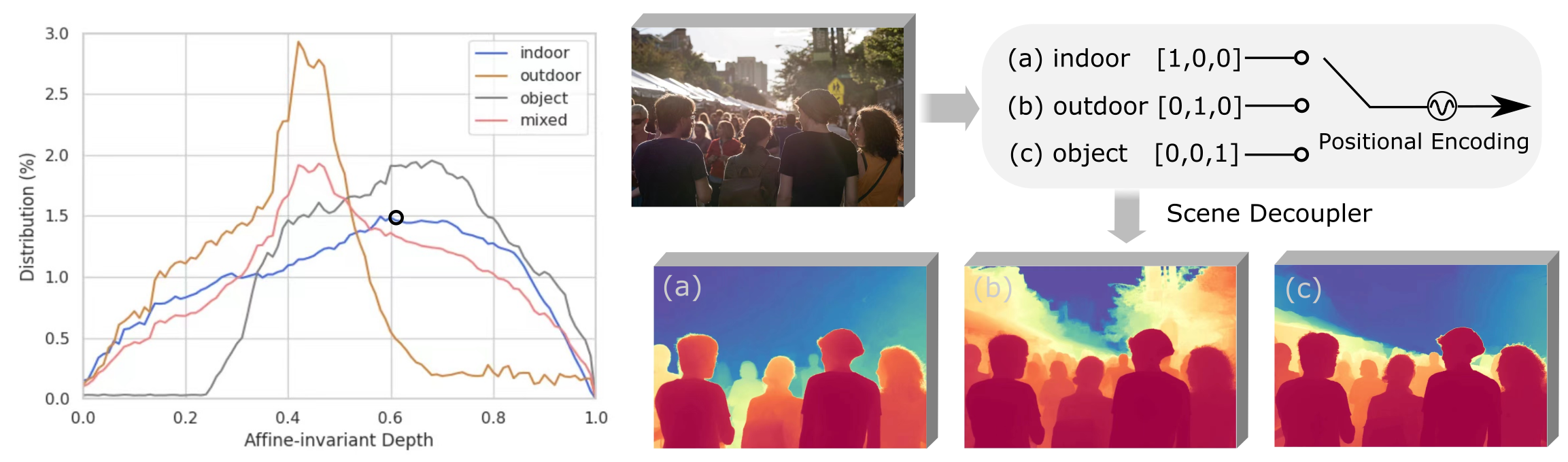}

\caption{\textbf{Scene Distributions (left) and Decoupler Structure as Guider (right).} We analyze the distributions of affine-invariant depth across three types of scenarios: indoor scenes, outdoor scenes, and background-free objects on our training dataset, where `mixed' refers to the mixture of the three types. 
To clarify, the black circle dot indicates that the proportion of affine-invariant depth in [0.595, 0.605] is 1.5\%. The Scene Decoupler encodes the one-hot domain vectors into positional embedding, which guides the stable diffusion to recognize the spatial layouts of different scene types.}
\vspace{-2em}
\label{fig:sceneswitcher}
\end{figure*}

\noindent
{\textbf{Scene Distribution Decoupler.}}
As we explore diverse scenarios, we encounter situations where the estimated geometry shows a bias towards unintended layouts, leading to significant compression of foreground elements. 
This occurs because stable diffusion models may struggle with figuring out the correct spatial layouts of the captured scenes due to the varied spatial structures depicted in the training data. 
For example, outdoor scenes often feature an infinite depth range, indoor scenes have a constrained depth range and background-free objects exhibit even narrower depth ranges.

A statistical analysis of scale-invariant depth distributions across different scene types is presented in~\figref{fig:sceneswitcher}, which shows that three types of scenes present different spatial structures.
If we adopt Gaussian distribution to model the spatial layouts, the depth distributions of the outdoor, indoor and object scenarios have different means and variances $(\mu_{1}, \sigma_{1}^{2})$, $(\mu_{2}, \sigma_{2}^{2})$ and $(\mu_{3}, \sigma_{3}^{2})$, respectively. 
The depth distribution of the mixed-up scenes tends to be a unified and neutralized distribution (red line) with $(\mu_{1}+\mu_{2}+\mu_{3}, \sigma_{1}^{2}+\sigma_{2}^{2}+\sigma_{3}^{2})$.
However, directly learning such a mixed distribution proves to be challenging.

To address the problem of layout ambiguity, we propose to learn the distinct three sub-distributions separately instead of directly learning the whole mixed distribution.
To achieve this, we introduce a Scene Distribution Decoupler to guide the diffusion model toward learning different distributions. 
Specifically, $(\hat{\bd}, \hat{\bn})=f(\bx, \bs_{\bi}), i = \{0,1,2\}$, where $\bs_{0}, \bs_{1}, \bs_{2}$ denote the one-hot vectors of indoor, outdoor, object scene types, respectively. Resembling geometry switcher, these one-dimensional vectors are processed by positional encoding and are then element-wisely added to the time embedding. 

\noindent
\textbf{{Loss Function.}} We adopt multi-resolution noises~\cite{Kasiopy2023multires,ke2023repurposing} to preserve low-frequency details in the depth and normal maps, as similar values will frequently appear in local geometric regions. This deviation proves to be more efficient than a single-scale noise schedule. We perturb the two geometry branches with the same time-step scheduler to decrease the difficulty when learning more modalities. Finally, we utilize the v-prediction~\cite{salimans2022progressive} as the learning objective:

\begin{equation}
\mathcal{L}=\mathbb{E}_{\mathbf{x}, \mathbf{d}, \mathbf{n}, \boldsymbol{\epsilon}, t, s} [\left\|\hat{\boldsymbol{\epsilon}}_{\boldsymbol{\theta}}\left(\bZ^{\bd}_{t} ; \bx, \bs_{\bd}, \bs_{i} \right)-\boldsymbol{\bv}^{\bd}_{t}\right\|_2^2+\left\|\hat{\boldsymbol{\epsilon}}_{\boldsymbol{\theta}}\left(\bZ^{\bn}_{t} ; \bx, \bs_{\bn}, \bs_{i} \right)-\boldsymbol{\bv}^{\bn}_{t}\right\|_2^2]
\end{equation}
where $\bv^{\bd}_{t}=\alpha_{t} \boldsymbol{\epsilon}^{\bd}_{t}-\sigma_{t}\bZ^{\bd}$ and $\bv^{\bn}_{t}=\alpha_{t} \boldsymbol{\epsilon}^{\bn}_{t}-\sigma_{t}\bZ^{\bn}$; $\boldsymbol{\epsilon}^{\bd}_{t}$ and $\boldsymbol{\epsilon}^{\bn}_{t}$ are two Gaussian noises independently sampled from multi-scale noise sets for depth and normal, respectively. The unified denoising network $\hat{\boldsymbol{\epsilon}}_{\boldsymbol{\theta}}$ with annealed noise scheduler generates the desired geometry noises conditioned by hierarchical switchers ($\bs_{\bd}, \bs_{\bn}, \bs_{i}$) and input image $\bx$.

\subsection{3D Reconstruction with Depth and Normal} \label{sec:3drecon}
With the estimated depth map $\hat{\bd}$ and normal map $\hat{\bn}$, we can reconstruct the underlying 3D structure based on the pinhole camera model. 
Since the predicted depth is affine-invariant with unknown scale and shift, it is not feasible to directly convert such a depth map into 3D point clouds with reasonable shapes.
To address it, we first optimize two parameters, i.e., scale $\hat{s}$ and shift $\hat{t}$ to formulate a metric depth map as $\hat{\bd} \times \hat{s} + \hat{t}$.
Then we calculate a 
normal map $\hat{\bn}_{\bd}$ by operating the least square fitting on depth~\cite{long2020occlusion}. We aim to minimize the difference between $\hat{\bn}_{\bd}$ and $\hat{\bn}$ to optimize $\hat{s}$ and $\hat{t}$. The objective function can be written as $\min _{\substack{\hat{s} , \hat{t}}} \text{D} (\hat{\bn}_{\bd}, \hat{\bn})$, where the normal difference is calculated in spherical coordinate.
With the optimized parameters scale and shift, we could obtain 
the ``pseudo'' metric depth, which is combined with the estimated normal map for surface reconstruction using the BiNI algorithm~\cite{cao2022bilateral}.

%% file: main/4_experiment.tex
\section{Experiment}

\subsection{Implementation Details and Datasets}

\noindent
\textbf{Implementation Details.}
We finetune the whole U-Net from the pre-trained Stable Diffusion V2 Model~\cite{rombach2022high}, which has been finetuned with image conditions. Our code is developed based on diffusers~\cite{patrick2022diffusers}. We use an image size of 576 × 768 and train the model for 20,000 steps with a total batch size of 256. This entire training procedure typically requires 2 days on a cluster of 8 Nvidia Tesla A100-40GB GPUs. We use the Adam optimizer with a learning rate of $1\times10^{-5}$. Additionally, to enhance dataset diversity, we apply random horizontal flipping,  crop, and photometric distortion (contrast, brightness, saturation, and hue) to the 2D image collection during training.

\noindent
\textbf{Training Datasets.}
We train our model on three categories: 1)Indoor: \textit{Hypersim}~\cite{roberts2021hypersim} is a photorealistic synthetic dataset with 461 indoor scenes. We filter out 191 scenes without tilt-shift photography. We further cull out incomplete images and finally obtain 25,463 samples. \textit{Replica}~\cite{straub2019replica} is a dataset of high-quality reconstructions of 18 indoor spaces. We filter out 50,884 samples with complete context. 2)Outdoor: \textit{3D Ken Burns}~\cite{niklaus20193d} provides a large-scale synthetic dataset with 76,048 stereo pairs in 23 in-the-wild scenes. We further incorporate 39,630 synthetic city samples in 1440$\times$3840 high resolutions from our own simulation platform. The normal GT is derived from the depth maps. (See Supp. for visualization) 3)Background-free Object: \textit{Objaverse}~\cite{deitke2023objaverse,qiu2023richdreamer} is a massive dataset of over 10 million 3D objects. We filter out 85,997 high-quality objects as training data.

\subsection{Evaluation}

\noindent
\textbf{Evaluation Datasets.}
We assess our model's efficacy across six zero-shot relative depth benchmarks, including NYUv2~\cite{silberman2012indoor}, KITTI~\cite{Geiger2013IJRR}, ETH3D~\cite{schops2017multi}, ScanNet~\cite{dai2017scannet}, DIODE~\cite{vasiljevic2019diode}, and OmniObject3D~\cite{wu2023omniobject3d}. For surface normal estimation, we employ in-total five benchmarks on NYUv2~\cite{silberman2012indoor, qi2018geonet}, ScanNet~\cite{dai2017scannet, huang2019framenet}, iBim-1~\cite{koch2018evaluation, bae2022irondepth}, DIODE-outdoor \cite{vasiljevic2019diode}, and OmniObject3D~\cite{wu2023omniobject3d} for zero-shot evaluation. 

\noindent
\textbf{Baselines.}
For affine-invariant depth estimation, we select baselines from state-of-the-art methods that demonstrate generalizability through training on diverse datasets. These methods are specialized in predicting either depth (DiverseDepth~\cite{yin2020diversedepth}, LeReS~\cite{yin2021learning}, HDN~\cite{zhang2022hierarchical}, Marigold~\cite{ke2023repurposing}) or disparity (MiDaS~\cite{Ranftl2022}, DPT~\cite{ranftl2021vision}, Omnidata~\cite{eftekhar2021omnidata}). For surface normal estimation, the field has seen fewer works~\cite{eftekhar2021omnidata, zhang2023jointnet, kar20223d} addressing zero-shot estimation specifically. Hereby, We choose both SoTA in-domain (EENSU~\cite{bae2021estimating}) and zero-shot methods (Omnidata v1~\cite{eftekhar2021omnidata}, v2~\cite{kar20223d}, and the ultra-recent DSINE~\cite{bae2024dsine}) as the baselines.

\subsubsection{Metrics.}
Building upon prior research~\cite{yin2023metric3d}, we assess the performance of depth estimation methods using the absolute relative error (AbsRel) and accuracy within a threshold $\delta^{\mathbf{1}}=1.25$. For surface normal estimation, we evaluate using the Mean angular error and accuracy within $11.25^\circ$, aligning with established methods~\cite{bae2021estimating}. We evaluate Geometric Consistency (GC) between depth and normal as follows: we first estimate the pseudo scale and shift of the estimated depth using GT depth, and then convert the estimated depth into metric depth.
We calculate the Mean angular error of the normal difference between predicted normal and normal calculated from the metric depth to evaluate the consistency between estimated depth and normal. 

\subsection{Comparison}

\begin{figure*}[h]
\centering
\includegraphics[width=\linewidth]{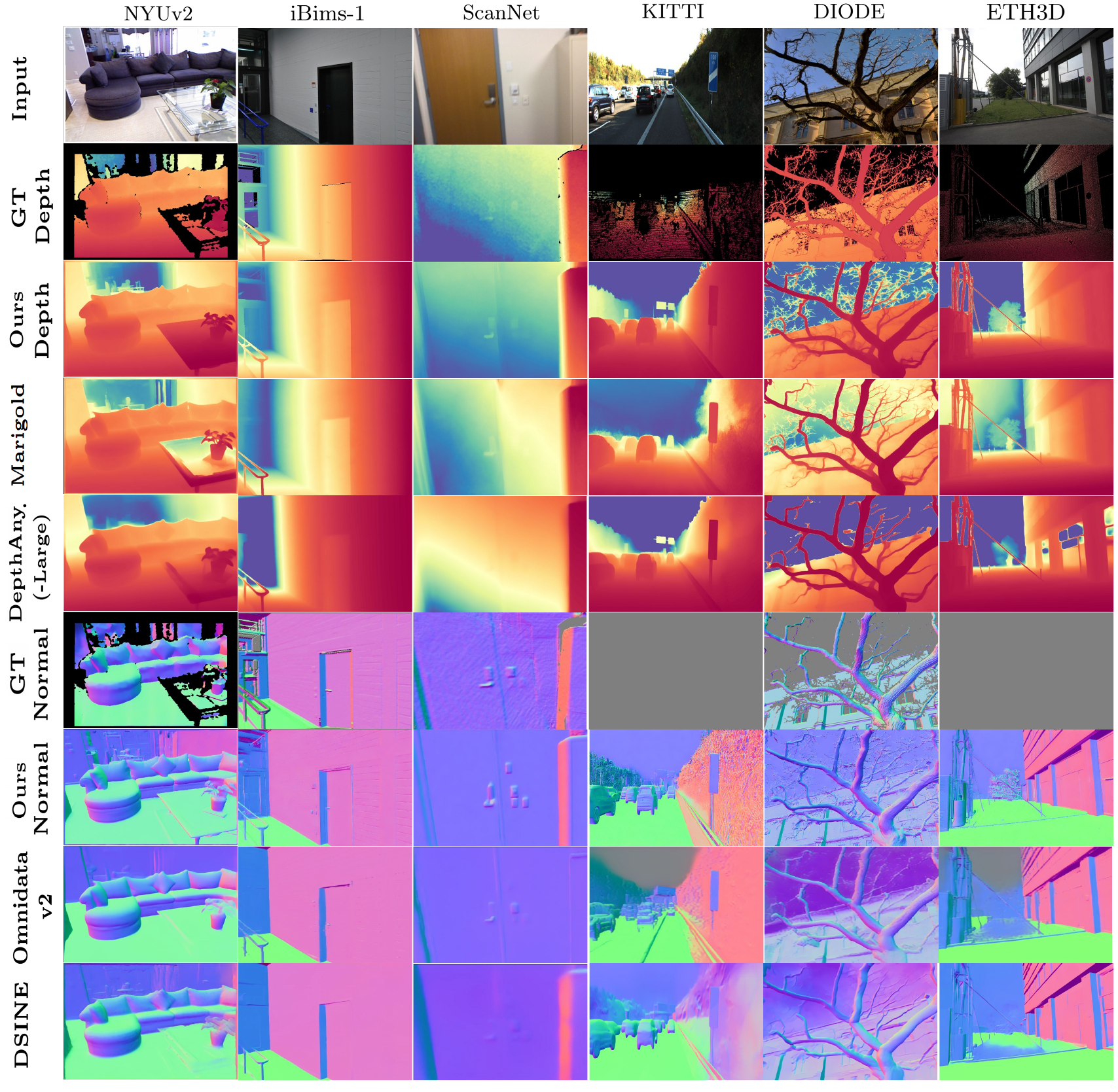}
\caption{Qualitative comparison on zero-shot depth and normal benchmarks.}
\label{fig: qual_dataset.}
\vspace{-2ex}
\end{figure*}

\input{tabs/depth_benchmark}

\noindent
\textbf{Depth Estimation.} We present the quantitative evaluations of zero-shot affine-invariant depth in~\tabref{tab: benchmark}. DepthAnything~\cite{yang2024depth} achieves the best quantitative numbers across three real datasets but presents a significant performance drop on unreal images (see~\figref{fig: qual_dataset.} and \figref{fig:inthewildcomp}). This may be because although DepthAnything is trained on 63.5M images, its discriminative nature limits its ability to generalize on images that significantly differ from training images. On the other hand, its results fail to capture rich geometric details. Compared to the robust depth estimator Marigold~\cite{ke2023repurposing}, GeoWizard shows more correct foreground-background relationships, especially in outdoor scenarios.

\input{tabs/normal_benchmark}

\subsubsection{Normal Estimation.}
\begin{figure*}[h]
 \centering
 \newcommand{\mywidth}{.98\textwidth}
 \setlength\tabcolsep{0.05em}
 \newcolumntype{P}[1]{>{\centering\arraybackslash}m{#1}}
 \def\arraystretch{0.50}
  \begin{tabular}{P{0.5em}P{0.5em} P{\mywidth}}
    \rot{\textbf{\tiny{Input}}} && \includegraphics[width=\mywidth]{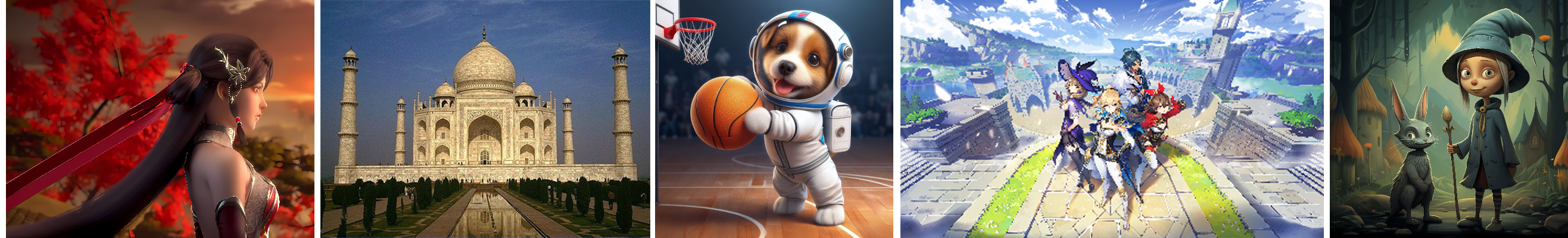} \\
     \rot{\textbf{\tiny{Ours Depth}}} && \includegraphics[width=\mywidth]{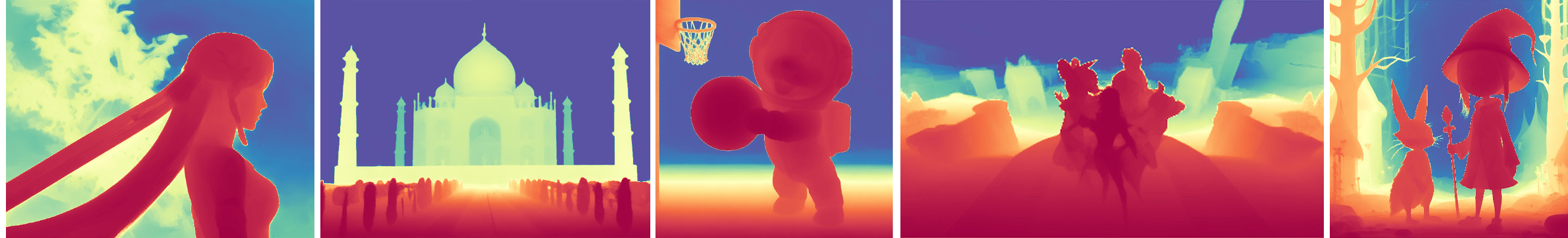}\\
     \rot{\textbf{\tiny{DepthAnything}}}&\rot{\textbf{\tiny{(-Large)~\cite{yang2024depth}}}}&\includegraphics[width=\mywidth]{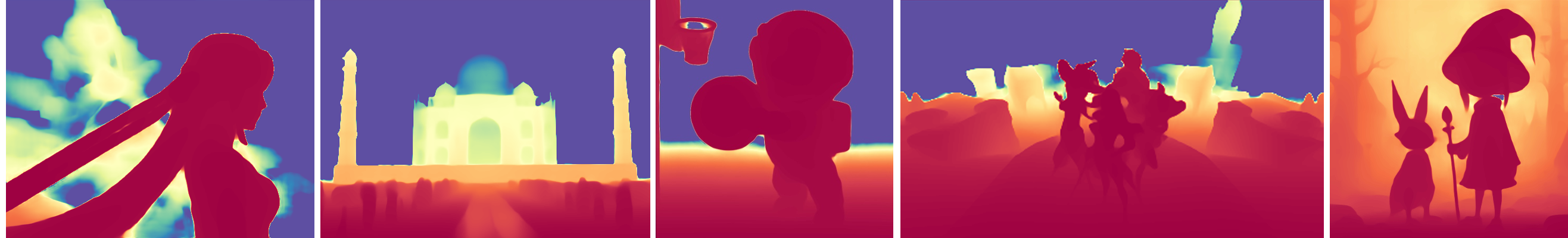} \\
     \rot{\textbf{\tiny{Ours Normal}}}& &\includegraphics[width=\mywidth]{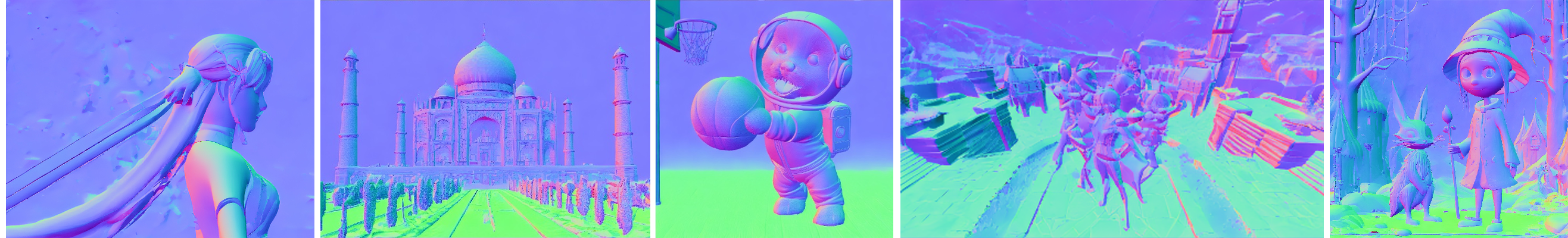} \\ \rot{\textbf{\tiny{DSINE}}}&\rot{\textbf{\tiny{\cite{bae2024dsine}}}}&\includegraphics[width=\mywidth]{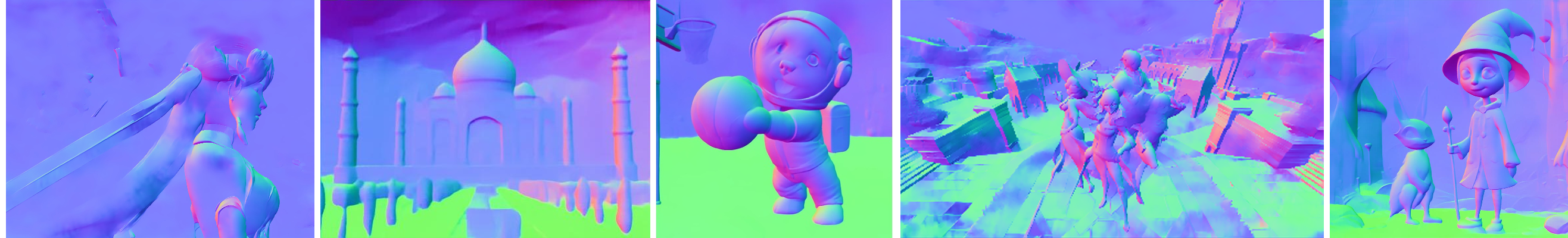} \\
 \end{tabular}
\caption{Geometry comparison on in-the-wild collections.
As discriminative models, DepthAnything and DSINE show significant performance drop on in-the-wild images, especially for the unreal images that are barely included in the collected training datasets. Please check more examples in the supplementary materials.}
\label{fig:inthewildcomp}
\vspace{-2em}
\end{figure*}

We present the quantitative evaluations of surface normal in~\tabref{tab: normal_benchmark}, where our method achieves superior performance. When compared with the SoTA normal approach DSINE~\cite{bae2024dsine}, our method recovers much finer-grained details and is more robust to unseen terrain in the~\figref{fig: qual_dataset.}. We further provide more out-of-domain comparisons in~\figref{fig:inthewildcomp}, where \textit{GeoWizard} surprisingly generates astonishing details and correct spatial structures. DSINE~\cite{bae2024dsine} can recover rough shape, but it struggles to produce high-frequency details, such as hairline, architectural texture, and limbs.

\begin{figure*}[h]
\centering
\includegraphics[width=\linewidth]{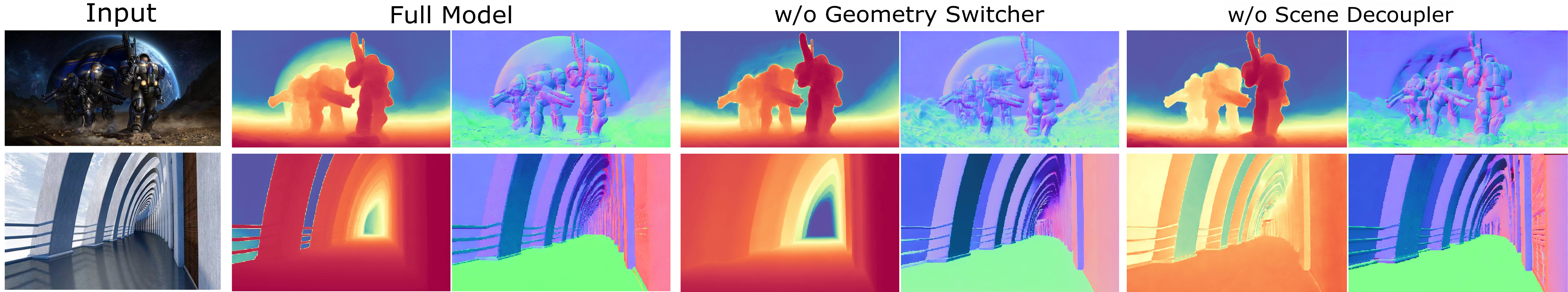}
\caption{Qualitative ablation. The geometric consistency decreases a lot, especially in far regions, when removing the cross-domain geometry switcher. 
Without our proposed Distribution Decoupler, the estimated depth and normal are mistakenly perceive the spatial layouts of the input images, like the Earth in the first row and the Sky in the second row.
}
\label{fig:ablation}
\vspace{-1em}
\end{figure*}

\input{tabs/ablation}

\subsection{Ablation Study}
We collect zero-shot validation sets that incorporate depth and normal from three scene distributions - the official test split of NYUv2~\cite{silberman2012indoor}, consisting of 654 images, and 138 high-quality samples from ScanNet~\cite{dai2017scannet} for indoor domain; the 432 in-the-wild samples from our simulation platform and filtered 86 images from DIODE~\cite{vasiljevic2019diode} for outdoor domain; 300 randomly selected real-world samples (over 40 categories) of OmniObject3D~\cite{wu2023omniobject3d} for object domain.

\noindent
\textbf{Joint Depth and Normal Estimation.}
We first investigate the effect of the geometry switcher. When removing the cross-domain geometry switcher (w/o Geometry Switcher), the overall geometric consistency drops significantly (16.2$\rightarrow$18.1, also as illustrated in \figref{fig:ablation}), verifying that cross-domain self-attention effectively correlates the two representations. 
We also train two diffusion models to separately learn depth and normal (Separate models), but this significantly reduces the performance across all evaluated metrics.

\noindent
\textbf{Decoupling Scene Distributions.} As we decouple the complex scene distribution into several sub-domains, GeoWizard can concentrate on a specific domain during in-the-wild inference. Therefore, it is not surprising that removing the decouple (w/o Scene Decoupler) leads to a performance drop across all domains (visually shown in \figref{fig:ablation}). 
Interestingly, the impact on the object domain is minimal, suggesting that object-level distribution is simpler to learn. 

\subsection{Application}
GeoWizard enables a wide range of downstream applications, including 3D reconstruction, novel view synthesis, and 2D content creation.

\begin{figure*}[h]
 \centering
 \newcommand{\mywidth}{.98\textwidth}
 \setlength\tabcolsep{0.05em}
 \newcolumntype{P}[1]{>{\centering\arraybackslash}m{#1}}
 \def\arraystretch{0.50}
  \begin{tabular}{P{0.5em}P{0.5em} P{\mywidth}}
    \rot{\textbf{\scriptsize{Input}}} && \includegraphics[width=\mywidth]{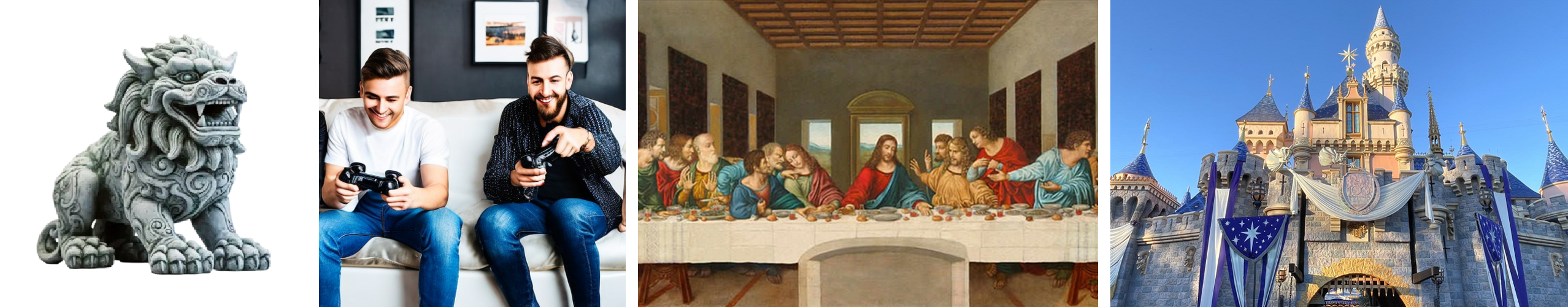} \\
     \rot{\textbf{\scriptsize{Ours}}} && \includegraphics[width=\mywidth]{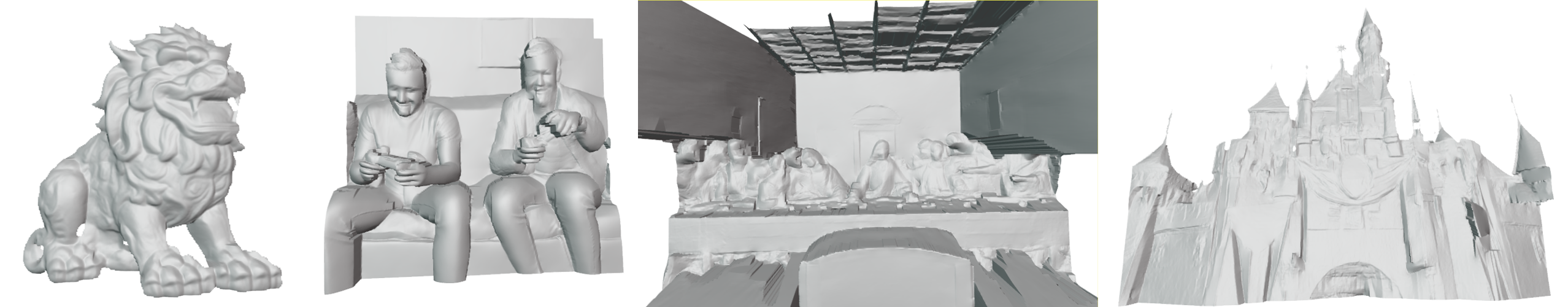}\\
     \rot{\textbf{\scriptsize{Omnidata}}}&\rot{\textbf{\scriptsize{v2~\cite{kar20223d}}}}& \includegraphics[width=\mywidth]{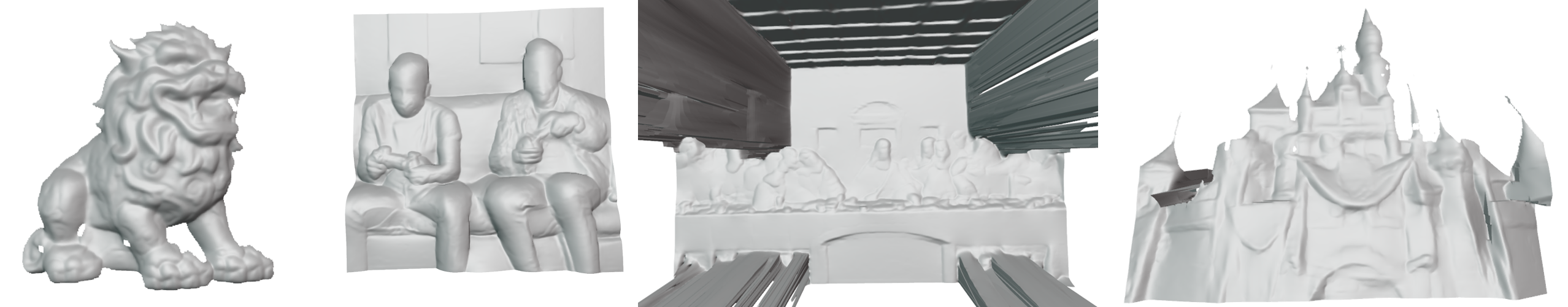}\\
 \end{tabular}
\caption{Geometry comparison on different scene domains. We relate the monocular depth and normal for directly recovering the 3D geometry. Ours consistently achieves more fine-grained details and spatial structure over Omnidata v2.}
\label{fig:surface_recon}
\end{figure*}

\input{tabs/monosdf}

\noindent
\textbf{3D Reconstruction with Geometric Cues.}
We can leverage the monocular geometric cues for surface reconstruction. With the algorithm outlined in \secref{sec:3drecon}, we can extract the 3D mesh directly. As depicted in~\figref{fig:surface_recon}, compared to Omnidata v2~\cite{kar20223d}, GeoWizard consistently generates finer details with higher fidelity and frequency detail (See the beard of the stone lion, the two men's faces, and castle sculpture) and more accurate 3D spatial structure (see the Last Supper). Additionally, we can condition these geometric cues to help surface reconstruction method~\cite{yu2022monosdf} to generate high-quality geometry. We conduct experiments on 4 scenes from ScanNet and employ evaluation metrics following~\cite{lyu2023learning,yu2022monosdf,guo2022neural}. \tabref{tab:monosdf} illustrates that our geometric guidance surpasses previous methods, particularly in terms of ``Recall'' and ``F-score''.

\begin{figure}[h]  
\centering    
\begin{subfigure}[b]{0.32\textwidth}  
    \centering  
    \includegraphics[width=\textwidth]{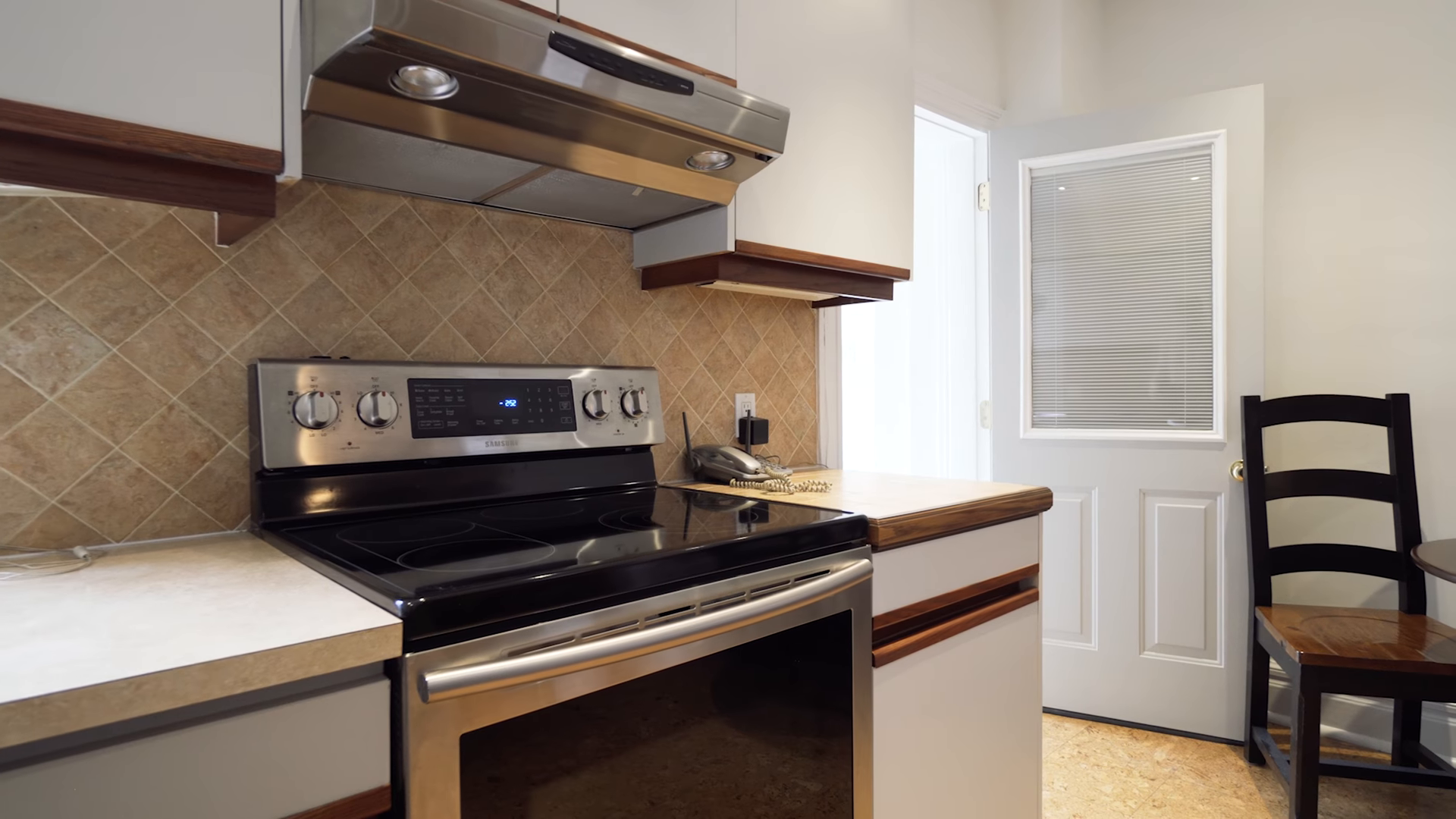}  
    \caption{Input}  
\end{subfigure}  
\hfill  
\begin{subfigure}[b]{0.32\textwidth}  
    \centering  
    \includegraphics[width=\textwidth]{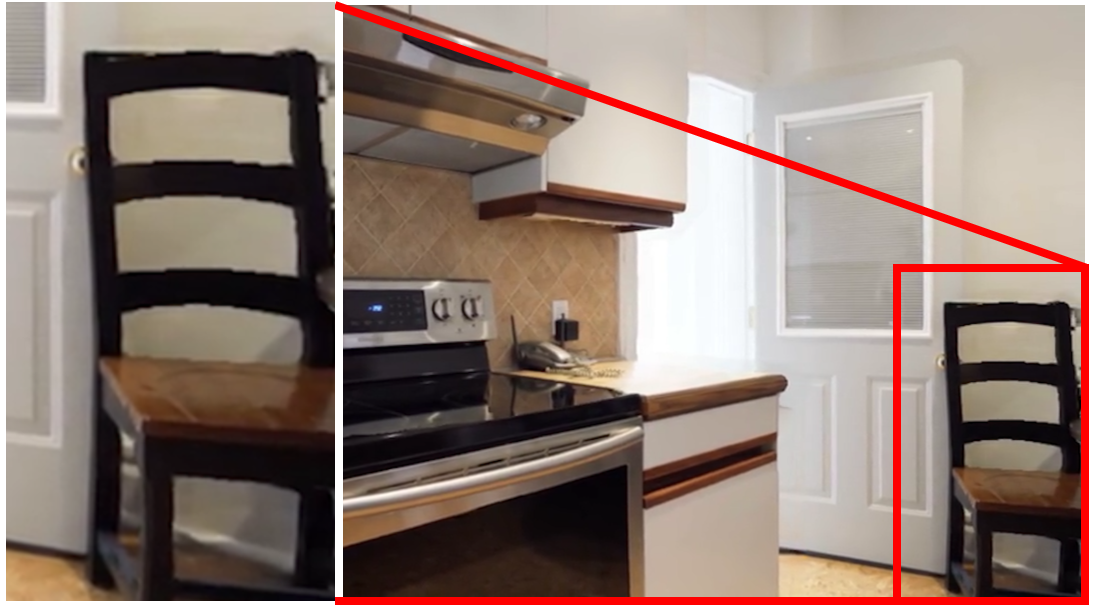}  
    \caption{GeoWizard (Ours)}  
\end{subfigure}  
\hfill  
\begin{subfigure}[b]{0.32\textwidth}  
    \centering  
    \includegraphics[width=\textwidth]{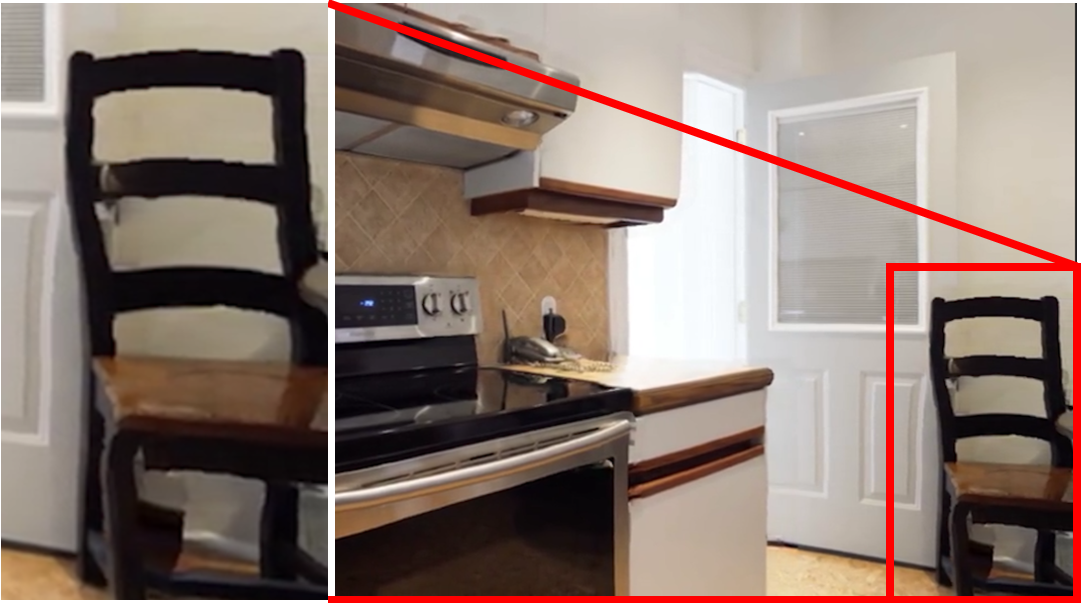}  
    \caption{Midas V3.1~\cite{Ranftl2022}}  
\end{subfigure}  
  
\caption{Novel view synthesis comparison. GeoWizard guides the~\cite{shih20203d} to generate more coherent and plausible structures like the thin chair legs and doorways.}  
\label{fig:novel view synthesis} 
\vspace{-2em}
\end{figure}  

\noindent
\textbf{Depth-aware Novel View Synthesis.}
We can utilize the depth cue generated by our model to enhance depth-based inpainting methods~\cite{shih20203d}. As shown in~\figref{fig:novel view synthesis}, compared to Midas V3.1~\cite{Ranftl2022}, GeoWizard achieves better novel view synthesis results (See the zoom-in chair within the red boxes) and enables more realistic 3D photography effect.

\begin{figure*}[h]
\centering
\includegraphics[width=\linewidth]{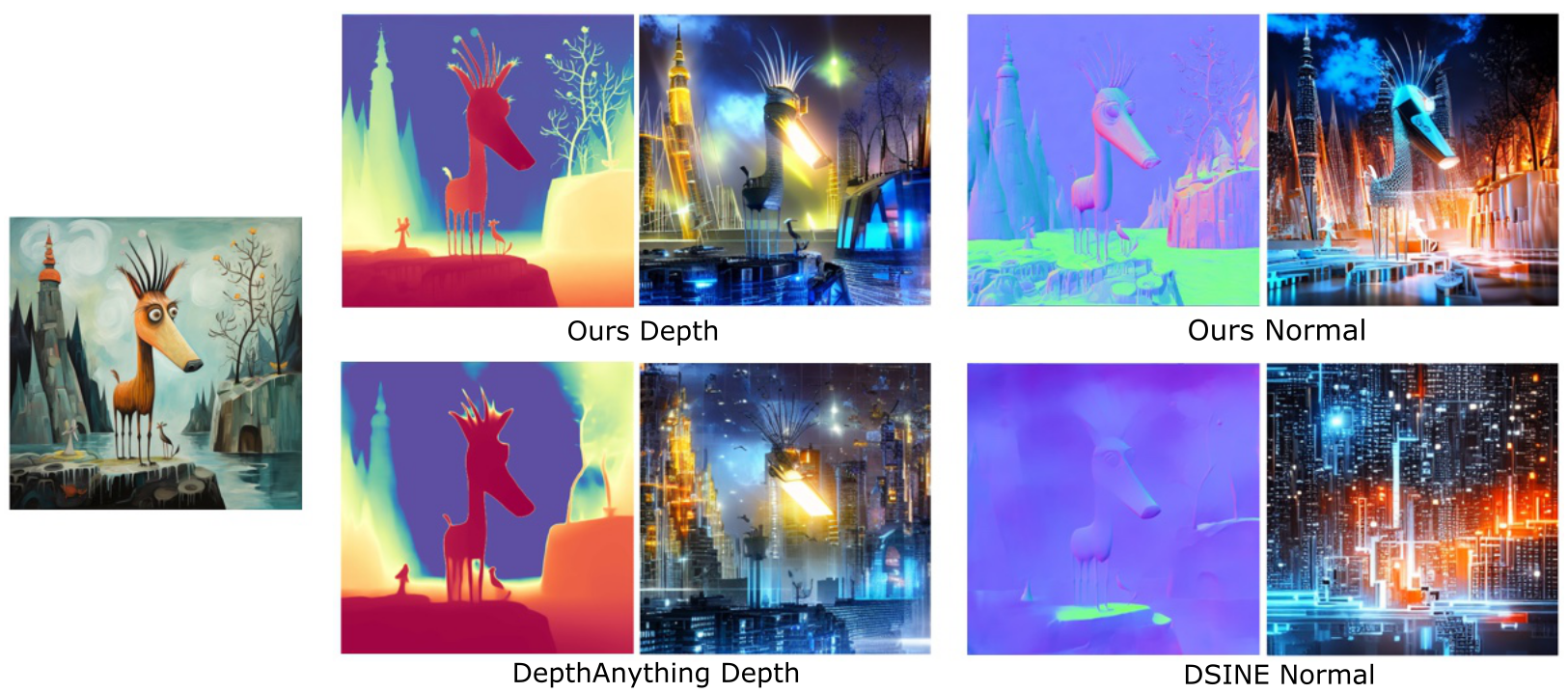}
\caption{Images generated by ControlNet conditioned on estimated depth maps and normal maps using text prompt \textit{``futuristic technology''}.}
\label{fig:controlnet}
\vspace{-2em}
\end{figure*}

\noindent
\textbf{2D Content Generation.}
We adopt depth/normal conditioned ControlNet~\cite{zhang2023adding} (SD 1.5) that takes spatial structure as input to evaluate the geometry indirectly. As depicted in~\figref{fig:controlnet}, the generated color images conditioned by our depth and normal are more semantically coherent to the original input image. However, the generated images conditioned on depth map of DepthAnything~\cite{yang2024depth} and normal map of DSINE~\cite{bae2024dsine} fail to keep similar 3D structures with the input image.

%% file: tabs/depth_benchmark.tex
\begin{table}[tb!]
\centering
\resizebox{\textwidth}{!} 
{
\begin{tabular}{c|cccccccccccc}
\toprule
\multirow{2}{*}{Method} &  \multicolumn{2}{c}{ NYUv2 } & \multicolumn{2}{c}{ KITTI } & \multicolumn{2}{c}{ ETH3D } & \multicolumn{2}{c}{ ScanNet } & \multicolumn{2}{c}{ DIODE-Full} & \multicolumn{2}{c}{ OmniObject3D} \\
& AbsRel $\downarrow$ & $\delta 1 \uparrow$ & AbsRel $\downarrow$ & $\delta 1 \uparrow$ & AbsRel $\downarrow$ & $\delta 1 \uparrow$ & AbsRel $\downarrow$ & $\delta 1 \uparrow$ & AbsRel $\downarrow$ & $\delta 1 \uparrow$ & AbsRel $\downarrow$ & $\delta 1 \uparrow$ \\

\midrule

\rowcolor{blue!10} DiverseDepth\cite{yin2020diversedepth} &  11.7 & 87.5 & 19.0 & 70.4 & 22.8 & 69.4 & 10.9 & 88.2 & 37.6 & 63.1 & - & - \\

\rowcolor{blue!10} MiDaS\cite{Ranftl2022} &  11.1 & 88.5 & 23.6 & 63.0 & 18.4 & 75.2 & 12.1 & 84.6 & 33.2 & 71.5 & - & - \\

\rowcolor{blue!10} LeReS\cite{yin2021learning} &  9.0 & 91.6 & 14.9 & 78.4 & 17.1 & 77.7 & 9.1 & 91.7 & 27.1 & 76.6 & - & - \\ 

\rowcolor{blue!10} Omnidata v2\cite{kar20223d} & 7.4 & 94.5 & 14.9 & 83.5 & 16.6 & 77.8 & 7.5 & 93.6 & 33.9 & 74.2  & 3.0 & \textbf{99.9} \\

\rowcolor{blue!10} HDN\cite{zhang2022hierarchical} & 6.9 & 94.8 & 11.5 & 86.7 & 12.1 & 83.3 & 8.0 & 93.9 & 24.6 & 78.0  & - & - \\

\rowcolor{blue!10} DPT\cite{ranftl2021vision} & 9.8 & 90.3 & 10.0 & 90.1 &  7.8 & 94.6 & 8.2 & 93.4 & \textbf{18.2} & 75.8   & - & - \\

\rowcolor{blue!10} Metric3D\cite{yin2023metric3d} & 5.8 & 96.3 & \textbf{5.8} & \textbf{97.0} &  6.6 & 96.0 & 7.4 & 94.1 & \underline{22.4} & 78.5   & - & - \\

\rowcolor{blue!10} DepthAnything\cite{yang2024depth} & \textbf{4.3} & \textbf{98.1} & 7.6 & 94.7 & 12.7 & 88.2 & \textbf{4.2} & \textbf{98.0} & 27.7 & 75.9 & \underline{1.8} & \textbf{99.9} \\

\midrule

\rowcolor{green!10} Marigold\cite{ke2023repurposing} &  5.5 & 96.4 & 9.9 & 91.6 & \underline{6.5} & \underline{96.0} & 6.4 & 95.1 & 30.8 & \underline{77.3}  & 3.0 & 99.8 \\

\rowcolor{green!10} GeoWizard (Ours) &  \underline{5.2} & \underline{96.6} & \underline{9.7} & \underline{92.1} & \textbf{6.4} &  \textbf{96.1} & \underline{6.1} & \underline{95.3} & 29.7 & \textbf{79.2} & \textbf{1.7} & \textbf{99.9} \\

\bottomrule

\end{tabular}
}
\caption{Quantitative comparison on 6 zero-shot affine-invariant depth benchmarks. We mark the best results in bold and the second best underlined. Discriminative methods are colored in \colorbox{blue!20}{blue} while generative ones in \colorbox{green!20}{green}. Please note that DepthAnything is trained on 63.5M images while ours is only trained on 0.28M images.}  

\vspace{-3em}
\label{tab: benchmark}
\end{table}

%% file: tabs/normal_benchmark.tex
\begin{table}[h]
\centering
\begin{threeparttable}
\resizebox{\textwidth}{!} 
{
\begin{tabular}{c|cccccccccc}
\toprule
\multirow{2}{*}{Method} & \multicolumn{2}{c}{ NYUv2 } & \multicolumn{2}{c}{ ScanNet } & \multicolumn{2}{c}{ iBims-1 } & \multicolumn{2}{c}{ DIODE-outdoor } & \multicolumn{2}{c}{OmniObject3D} \\
& Mean $\downarrow$ & $11.25^\circ \uparrow$ & Mean $\downarrow$ & $11.25^\circ  \uparrow$ & Mean $\downarrow$ & $11.25^\circ \uparrow$ & Mean $\downarrow$ & $11.25^\circ \uparrow$ & Mean $\downarrow$ & $11.25^\circ \uparrow$ \\

\midrule

EESNU~\cite{bae2021estimating} & \textbf{16.2} & \underline{58.6} & - & - & 20.0  & 58.5 & 29.5 & 26.8  & 31.9 & 18.8  \\

Omnidata v1~\cite{eftekhar2021omnidata} & 23.1 & 45.8 & 22.9 & 47.4 & 19.0 & 62.1  & 22.4 & 38.4 & 23.1 & 42.6  \\

Omnidata v2~\cite{kar20223d} & 17.2 & 55.5 & \underline{16.2} & 60.2 & 18.2 & 63.9  & \underline{20.6} & \underline{40.6} & \underline{21.4} & \underline{46.1} \\

DSINE~\cite{bae2024dsine} & \underline{16.4} & \textbf{59.6} & \underline{16.2} & \underline{61.0} & \underline{17.1} & \textbf{67.4}  & \textbf{19.3} & \textbf{44.1}  & 21.7 & 45.1 \\

\midrule

GeoWizard (Ours)  & 17.0 & 56.5 & \textbf{15.4} & \textbf{61.6} & \textbf{13.0} & \underline{65.3}  &  \underline{20.6} &  38.9 & \textbf{20.8} & \textbf{47.8}   \\

\bottomrule

\end{tabular}
}
\begin{tablenotes}
\scriptsize
\item[-]: EENSU \cite{bae2021estimating} is trained on ScanNet, thus the in-domain performance is omitted.
\end{tablenotes}

\caption{Quantitative comparison across 5 zero-shot surface normal benchmarks. Top results are highlighted in bold while second-best are underlined.} 
\vspace{-4em}
\label{tab: normal_benchmark}
\end{threeparttable}
\end{table}

%% file: tabs/ablation.tex
\begin{table}[h]
\centering
\resizebox{\textwidth}{!} 
{
\begin{tabular}{c|ccc|ccc|ccc|ccc}
\toprule
\multirow{2}{*}{Method} & \multicolumn{3}{c}{ Indoor } & \multicolumn{3}{c}{Outdoor} & \multicolumn{3}{c}{Object} & \multicolumn{3}{c}{Overall} \\

& AbsRel $\downarrow$ & Mean $\downarrow$ & GC $\downarrow$ & AbsRel $\downarrow$ & Mean $\downarrow$ & GC $\downarrow$ & AbsRel $\downarrow$ & Mean $\downarrow$ & GC $\downarrow$ & AbsRel $\downarrow$ & Mean $\downarrow$ & GC $\downarrow$ \\

\midrule
Separate models  & 7.4 & 15.1 & 18.2 & 12.5 & 26.2 & 27.9 & 5.2 & 18.2 & 20.1 & 8.5 & 16.9 & 19.1 \\
w/o Geometry Switcher & 5.7 & 13.1 & 17.3 & 9.8 & 22.3 & 27.1 & \textbf{3.3} & 15.8 & 18.5 & 6.9 & 15.0 & 
18.1 \\
w/o Scene Decoupler & 5.8 & 13.8 & 15.4 & 10.5 & 24.7 & 24.5 & 3.7 & 15.5 & 17.9 & 7.5 & 16.1 & 16.5 \\

Full Model & \textbf{5.5} & \textbf{12.6} & 14.7 & \textbf{9.6} & \textbf{22.1} & \textbf{23.5} & 3.5  & \textbf{15.4} & \textbf{17.6} & \textbf{6.7} & \textbf{14.8} & \textbf{16.2} \\

\bottomrule
\end{tabular}
}

\caption{Quantitative ablation across three types of scenarios. Here we use AbsRel to evaluate depth accuracy, and Mean for normal accuracy. Furthermore, we also report the geometric consistency (GC) of the two representations.} 
\vspace{-3em}
\label{tab:ablation}
\end{table}

%% file: tabs/monosdf.tex
\begin{table}[h]
\centering
\resizebox{.65\textwidth}{!} 
{
\begin{tabular}{c|cccccc}
\toprule
Geometric Cues & Acc$\downarrow$ & Comp$\downarrow$ & C-$\mathcal{L}_{1}$ $\downarrow$ & Prec$\uparrow$ & Recall $\uparrow$ & F-score$\uparrow$ \\
\midrule
Omnidata v2~\cite{kar20223d} & 0.035 & 0.048 & 0.042 & 79.9 & 68.1 & 73.3  \\
DSINE~\cite{bae2024dsine} & 0.036 & 0.045 & 0.040 & \textbf{80.1} & 70.2 & 74.7 \\
\midrule
GeoWizard (Ours) & \textbf{0.033} & \textbf{0.042} & \textbf{0.038} & 80.0 &  \textbf{70.7} & \textbf{75.1} \\
\bottomrule
\end{tabular}
}

\caption{Assessments of different geometric guidance used for MonoSDF~\cite{yu2022monosdf} on the ScanNet~\cite{dai2017scannet} dataset.}

\vspace{-3em}
\label{tab:monosdf}

\end{table}

%% file: main/5_conclusion.tex
\section{Conclusion}
In this work, we present \textit{GeoWizard}, a holistic diffusion model for geometry estimation. We distill the rich knowledge in the pre-trained stable diffusion to boost the task of high-fidelity depth and normal estimation. 
Using the proposed geometry switcher, \textit{GeoWizard} jointly produces depth and normal using a single model. By decoupling the mixed and sophisticated distribution of all scenes into several distinct sub-distributions, our model could produce 3D geometry with correct spatial layouts for various scene types. In the future, we plan to decrease the number of denoising steps to speed up the inference of our method. The latent consistency models~\cite{luo2023latent} may be leveraged to train a few-step diffusion model so that the inference time may be decreased to less than 1 second. 

%% file: main/6_acknowledgement.tex
\section*{Acknowledgments}
We thanks for the experimental help from Yichong Lu, and valuable suggestions from Shangzhan Zhang and Yuwei Guo.

\vspace{-1em}

%% file: main/7_supp.tex
\setcounter{table}{0}
\setcounter{figure}{0}
\renewcommand{\thetable}{R\arabic{table}}
\renewcommand\thefigure{S\arabic{figure}}

\appendix

\section{Appendix}

In this supplementary, we offer more details on implementation and experiments in~\appref{supp:implement} and~\appref{supp:experiment}, respectively. We also include more qualitative comparisons regarding depth and normal on zero-shot test sets and in-the-wild scenarios, 3D reconstruction, and novel view synthesis in~\appref{supp:qualitative}. Finally, we discuss limitations and potential negative impact in~\appref{supp:limitation}.

\section{Implementation Details} \label{supp:implement}

\subsection{Data Preprocessing}
We standardize the resolution at 576$\times$768 to blend samples from various scene distributions. To maintain the original aspect ratio, we resize the shorter side of a sample to 576 and randomly crop along the longer side. In the data augmentation strategy, we assign photometric distortion probabilities of 0.05, 0.1, and 0.05, and greyization probabilities of 0.1, 0.2, and 0.1 for indoor, outdoor, and object level, respectively. We set the far plane to be 80 meters in both 3D Ken Burns~\cite{niklaus20193d} and our own city dataset for outdoor scenes, and 5 meters in Objaverse~\cite{deitke2023objaverse} for background-free objects. We also define the normal orientation in these far (background) regions along the z-axis. In the Replica dataset~\cite{straub2019replica}, we exclude samples with fewer than 50 invalid pixels, designating the invalid areas to represent distant depths and background normals.

\subsection{Our Synthetic Urban Dataset}
\begin{figure*}[h]
\centering
\includegraphics[width=\linewidth]{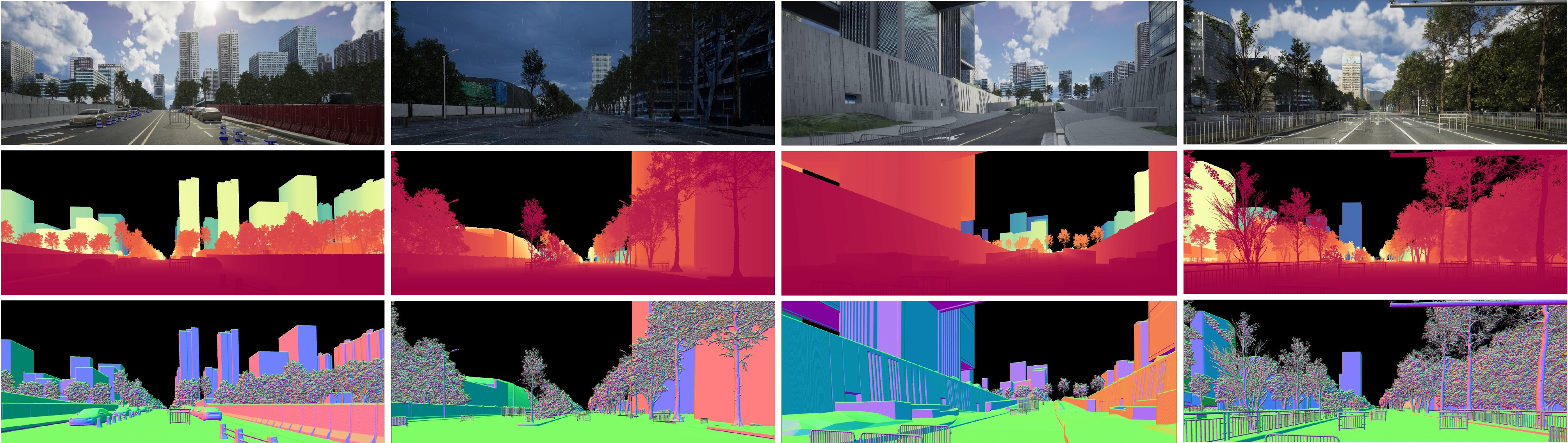}
\caption{Some samples of our rendered dataset. We mask the regions whose depth values are larger than 80m as black for better visualization.}
\label{fig:supp_simulation}
\vspace{-1em}
\end{figure*}

We first tried to add Virtual KITTI~\cite{cabon2020virtual} to involve more driving scenarios but ultimately decided against it, as the generated normal map is of low quality due to the limited resolution of depth map (375$\times$1242). As an alternative, we utilize Unreal Engine to create high-resolution (1440$\times$3840) urban samples (see~\figref{fig:supp_simulation}), and derive the normal map from depth using the least square algorithm. Our synthetic data encompasses a wide variety of city entities under different environmental conditions. Since the data is clean and complete, it allows our model to learn high-quality outdoor patterns.

\section{Experimental Details} \label{supp:experiment}

\subsection{Limitation on Normal GT}

\begin{figure*}[h]
\centering
\includegraphics[width=\linewidth]{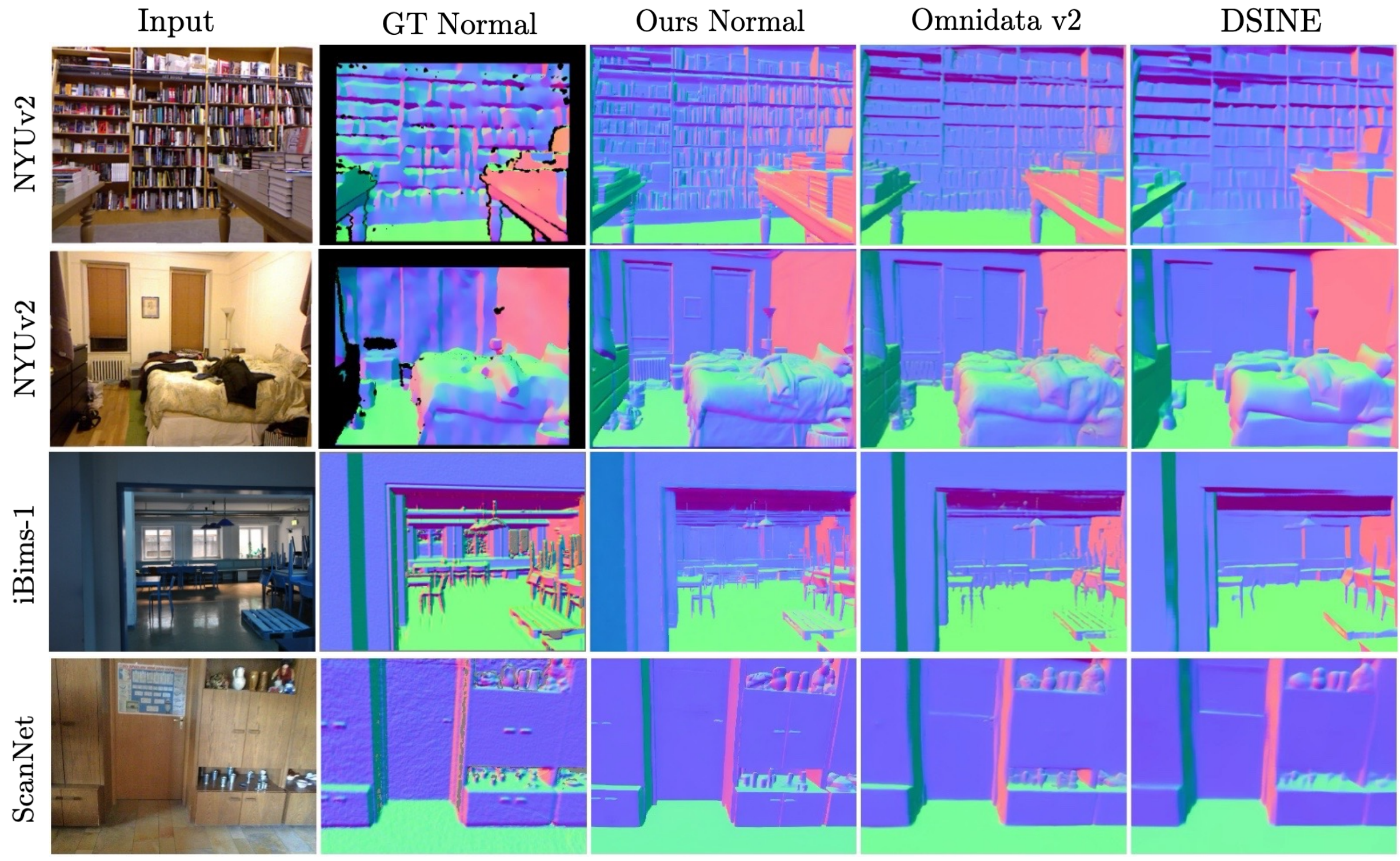}
\caption{Effect of noisy GT normal map. Our normal maps here display the best visual effect but are inferior in quantitative comparison with Omnidata v2 or DSINE.}
\label{fig:supp_noisynormalgt}
\end{figure*}

During our zero-shot tests on traditional normal benchmarks, we discovered that a lot of normal GT maps have noise, potentially impacting measurement precision. As shown in~\figref{fig:supp_noisynormalgt}, NYUv2's normal maps struggle with fine details such as book outlines, shelf edges, and folds, and even incorrectly represent the flat wall surfaces. Likewise, the normal maps from iBims-1 (limited resolution) and ScanNet (unexpected surface undulation and poor fine detail capture) are also of low quality. Thus, the quantitative comparisons presented in the main paper may only partially reflect the ground truth.
\subsection{More Ablation Studies}

\subsubsection{Applying Erroneous Domain Indicator}
When using the wrong domain-specific indicator for testing across various domains, we see a decline in both depth and normal (see~\tabref{tab:ablation_supp}), especially during zero-shot tests on indoor and outdoor benchmarks with an object indicator (w/ Object Indicator). This result makes sense since the indicator directs the model to concentrate on a specific distribution. We also observe that the geometric consistency seems to remain stable or even improved (14.7$\rightarrow$14.4 on indoor test with an outdoor indicator), suggesting the model's adaptability and robustness when guided by an out-of-domain indicator.

\subsubsection{Geometric Modeling}

We also study shared geometry embedding~\cite{liu2023hyperhuman} by increasing the dimension of the input in input (`w/ Shared Geometry' in~\tabref{tab:ablation_supp}). Without the specialized geometry switcher and using the same training iterations, we observe that this alternative converges more slowly, and the overall quality of depth and normal quality both decrease (6.7$\rightarrow$7.2, 14.8$\rightarrow$15.3), whereas the geometric consistency remains relatively unchanged. 

\input{tabs/ablation_supp}

\section{More Qualitative Comparisons} \label{supp:qualitative}
\subsection{Testset Depth and Normal}

\begin{figure*}[h]
\centering
\includegraphics[width=\linewidth]{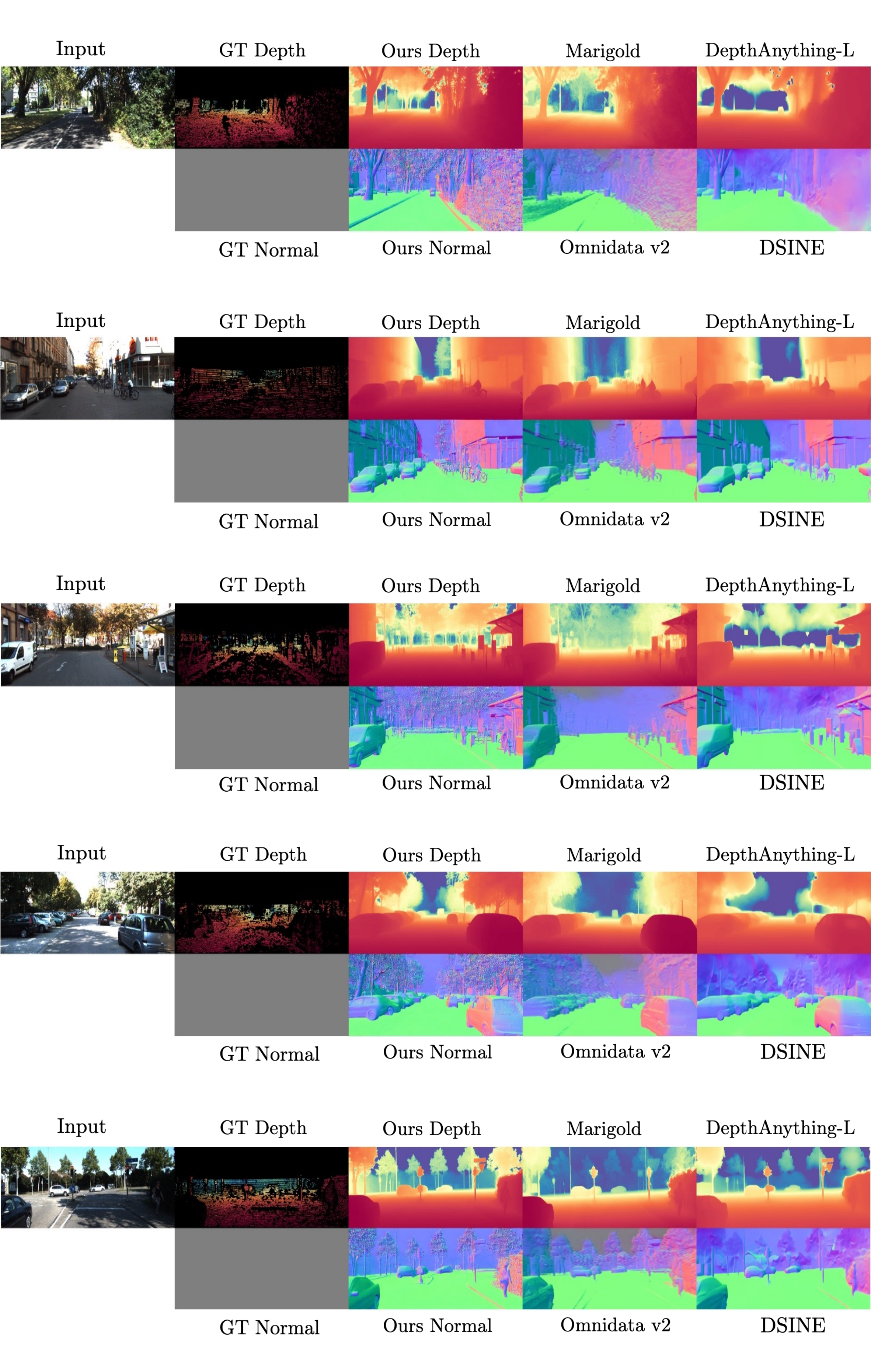}
\caption{Qualitative comparison on KITTI~\cite{Geiger2013IJRR}.}
\label{fig:supp_test_comp_kitti}
\end{figure*}

\begin{figure*}[h]
\centering
\includegraphics[width=\linewidth]{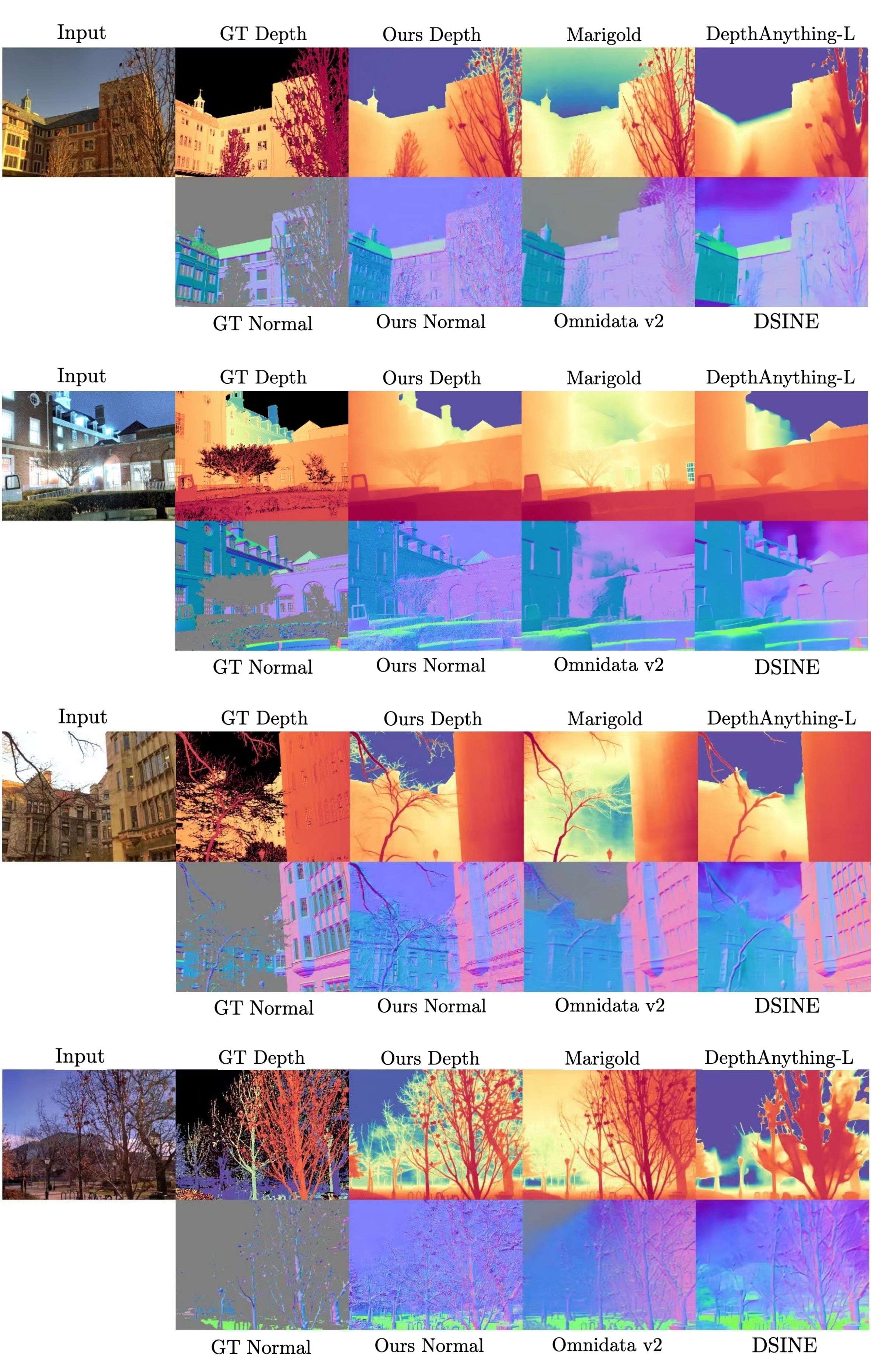}
\caption{Qualitative comparison on DIODIE~\cite{vasiljevic2019diode}.}
\end{figure*}

\begin{figure*}[h]
\centering
\includegraphics[width=\linewidth]{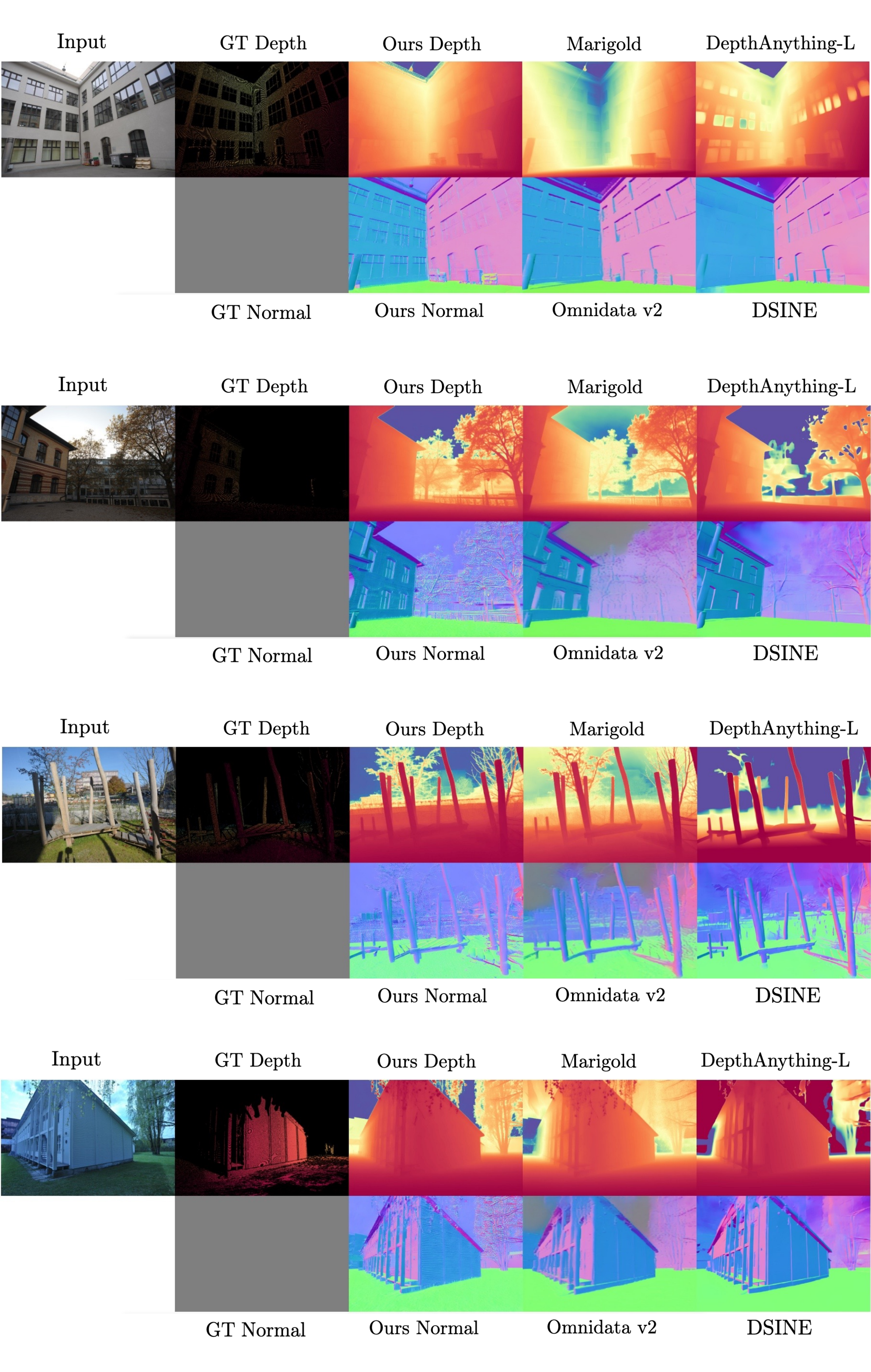}
\caption{Qualitative comparison on ETH3D~\cite{schops2017multi}.}
\end{figure*}

\begin{figure*}[h]
\centering
\includegraphics[width=\linewidth]{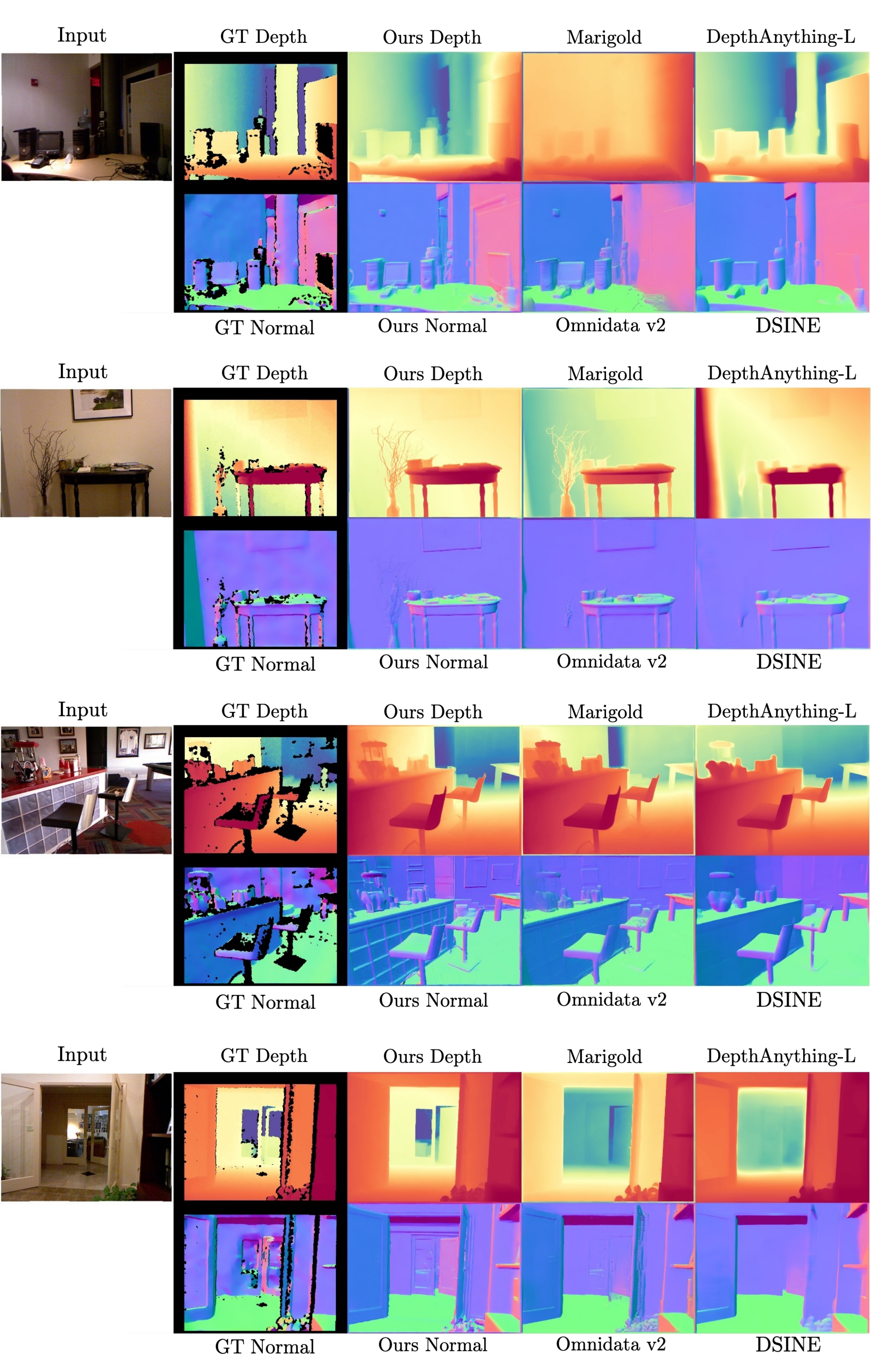}
\caption{Qualitative comparison on NYUv2~\cite{silberman2012indoor}.}
\end{figure*}

\begin{figure*}[h]
\centering
\includegraphics[width=\linewidth]{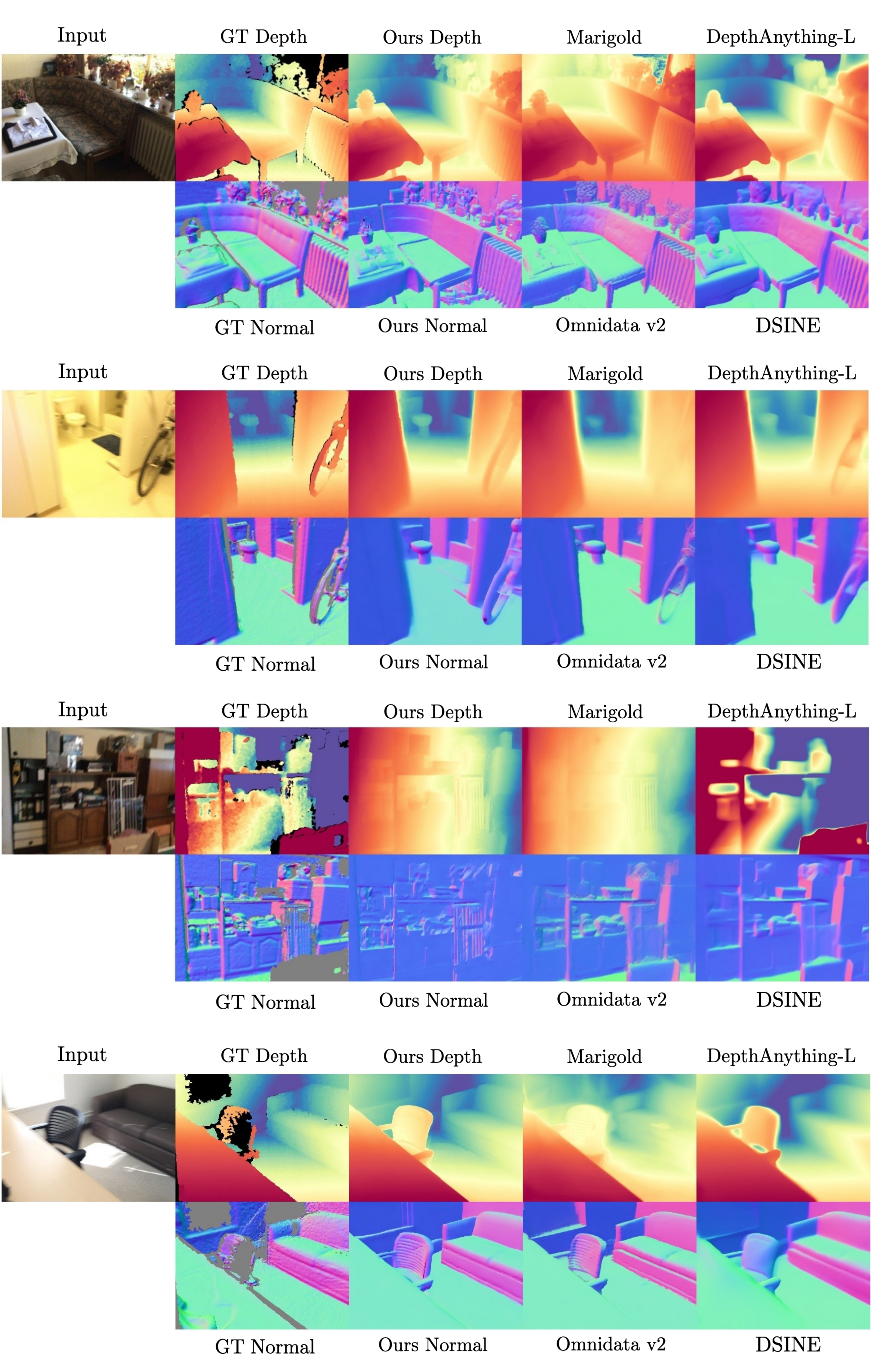}
\caption{Qualitative comparison on ScanNet~\cite{dai2017scannet}.}
\end{figure*}

\begin{figure*}[h]
\centering
\includegraphics[width=\linewidth]{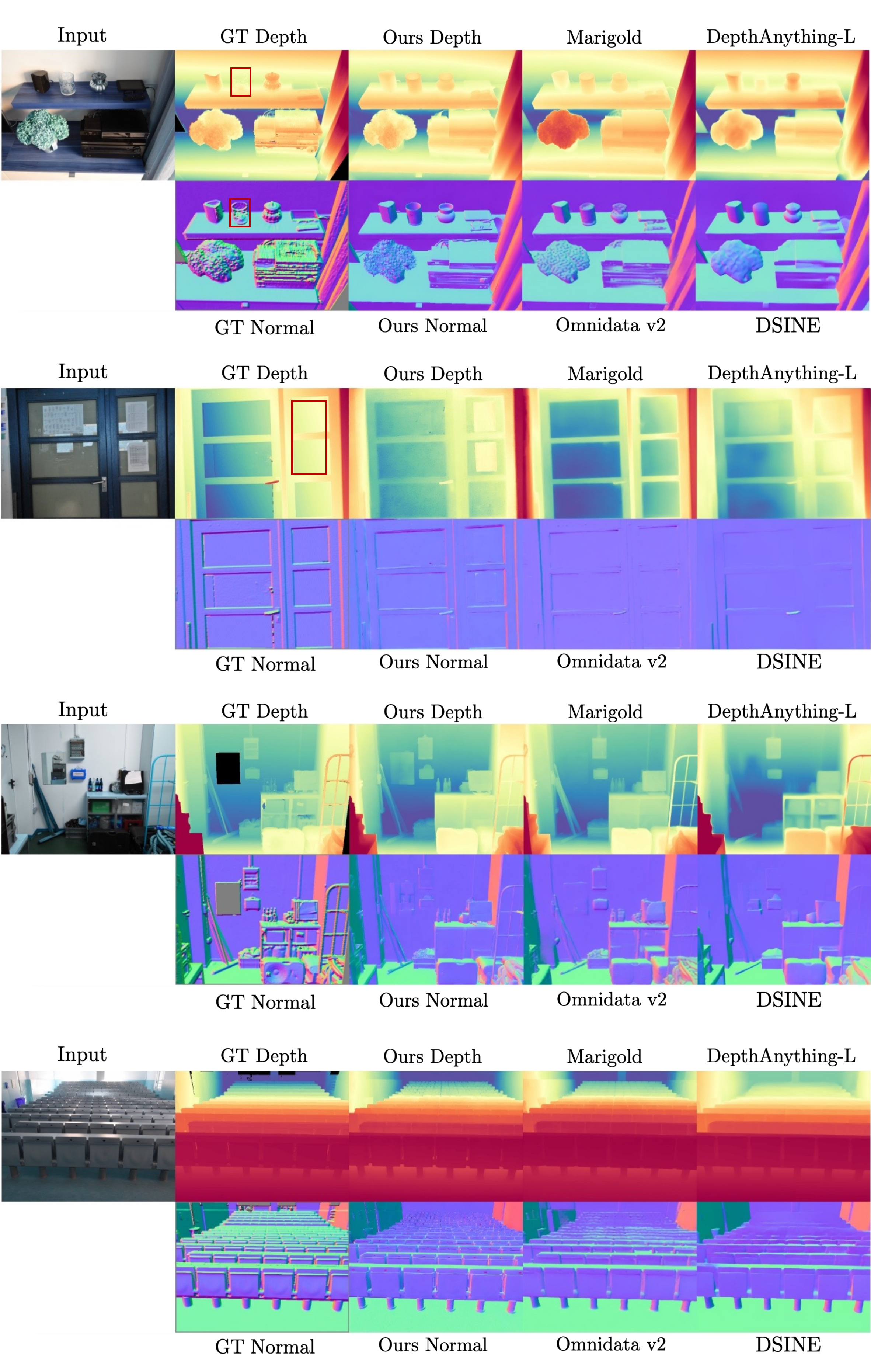}
\caption{Qualitative comparison on iBims-1~\cite{koch2018evaluation}. The red box marks the part where GT is erroneous.}
\end{figure*}

\begin{figure*}[h]
\centering
\includegraphics[width=\linewidth]{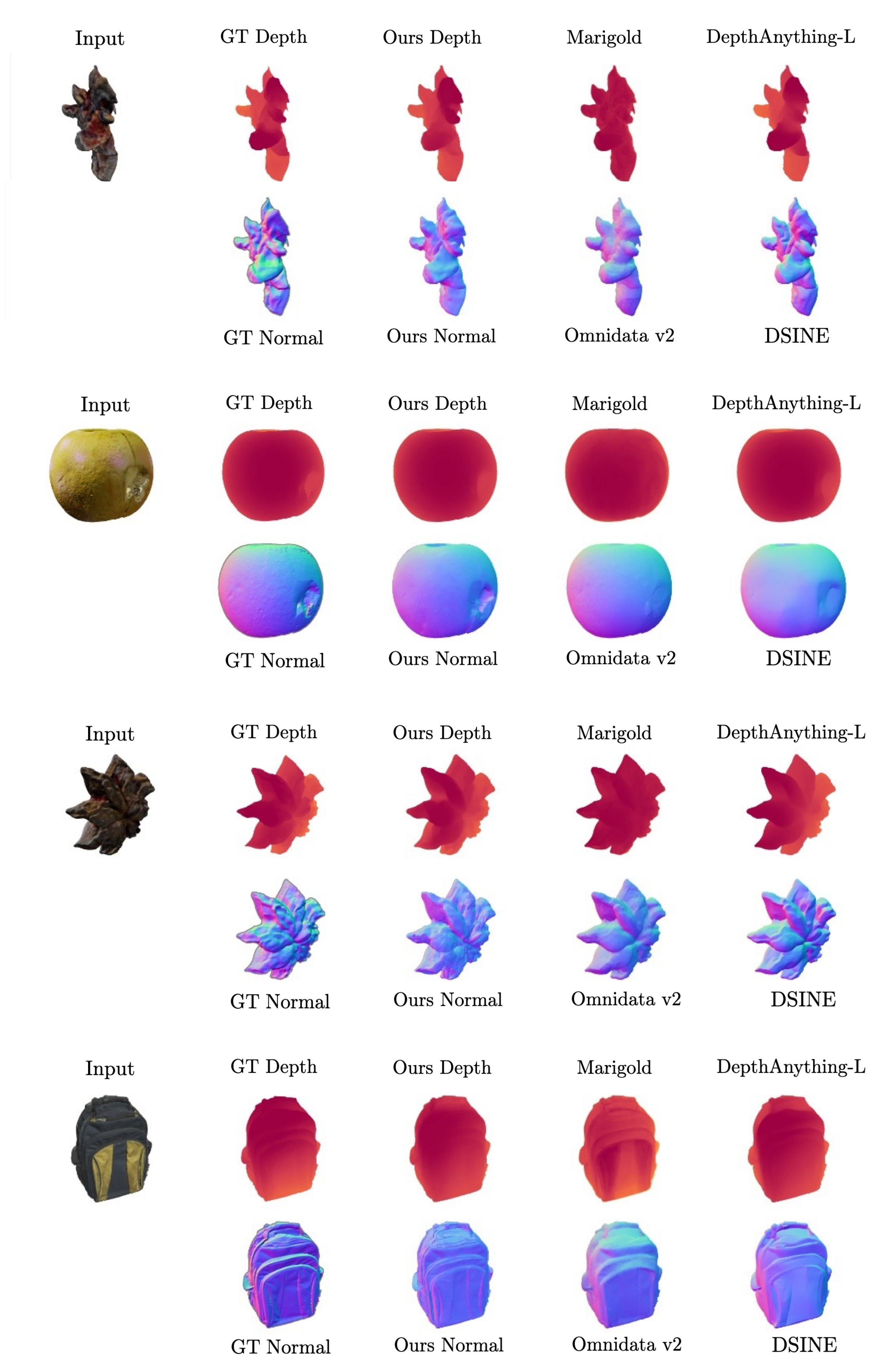}
\caption{Qualitative comparison on OmniObject3D~\cite{wu2023omniobject3d}.}
\label{fig:supp_test_comp_oo3d}
\end{figure*}

We include additional qualitative comparisons across 7 zero-shot test datasets\cite{silberman2012indoor,Geiger2013IJRR,schops2017multi,dai2017scannet,vasiljevic2019diode,wu2023omniobject3d,koch2018evaluation}, where our model is evaluate against Marigold~\cite{ke2023repurposing} and DepthAnything-L~\cite{yang2024depth} for depth, and agins Omnidata v2~\cite{kar20223d} and DSINE~\cite{bae2024dsine} for normal. These comparisons, visualized in~\figref{fig:supp_test_comp_kitti} to~\figref{fig:supp_test_comp_oo3d}, cover both depth and normal maps. To enhance visual contrast, we initially math the inverse relative depth from DepthAnything with the inverse GT depth. Following this affine alignment, we further convert it into actual depth. Note that the GT normal maps are shown in default grey when unavailable. For outdoor scenes, the `sky' in our normal maps is colored in pure blue [0,0,255] to denote the standard orientation [0,0,1]. In comparison on iBim-1, we mask out the erroneous parts in GT with red boxes. Overall, Geowizard consistently outperforms in generating detailed high-frequency details across all datasets, although the difference might not be as discernible in OmniObject3D due to its simplistic object structures.

\subsection{In-the-Wild Depth and Normal}
\begin{figure*}[h]
\centering
\includegraphics[width=.98\linewidth]{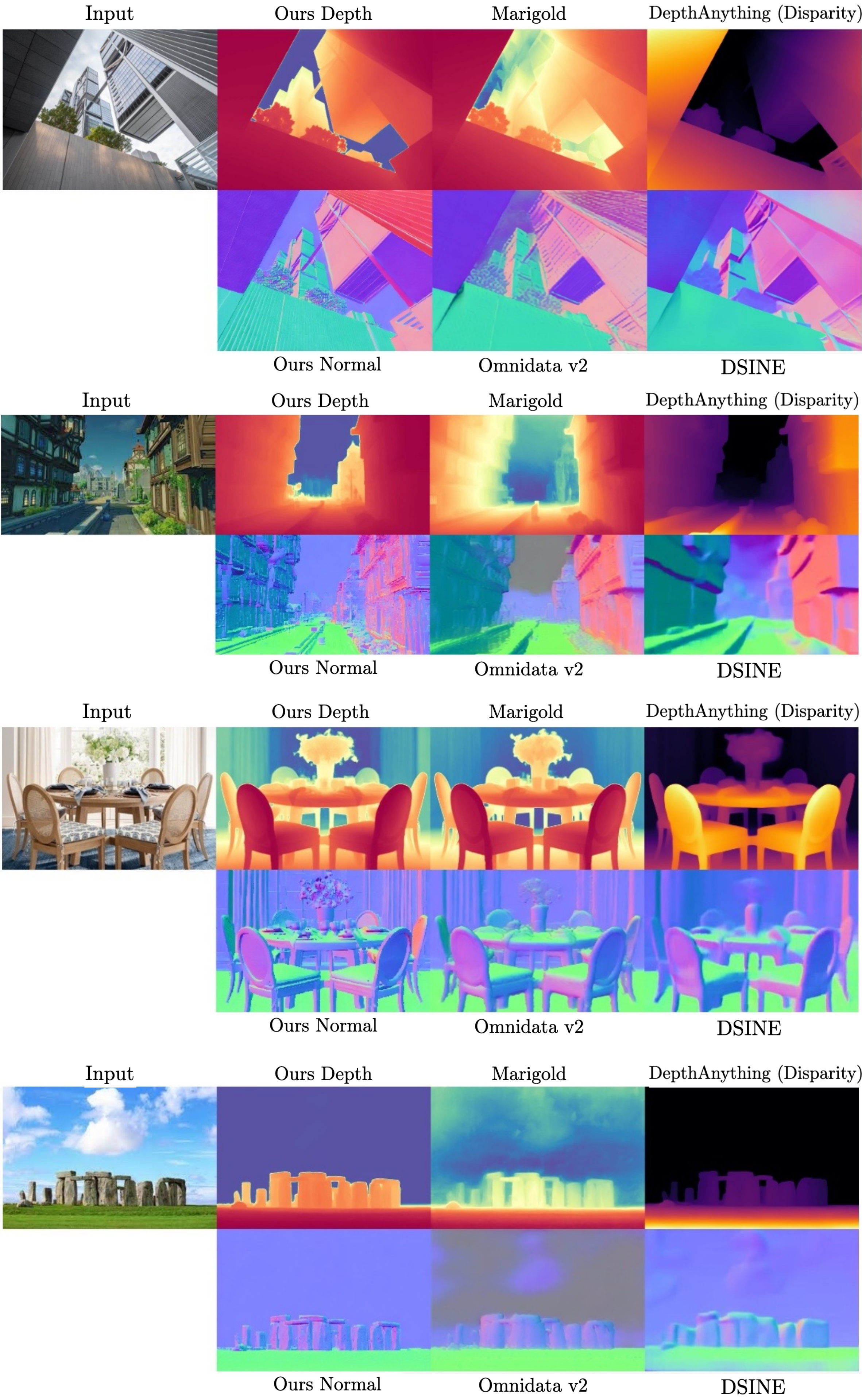}
\caption{Qualitative geometry comparison on in-the-wild images (1/11).}
\label{fig:inthewild_comp_1}
\end{figure*}

\begin{figure*}[h]
\centering
\includegraphics[width=.98\linewidth]{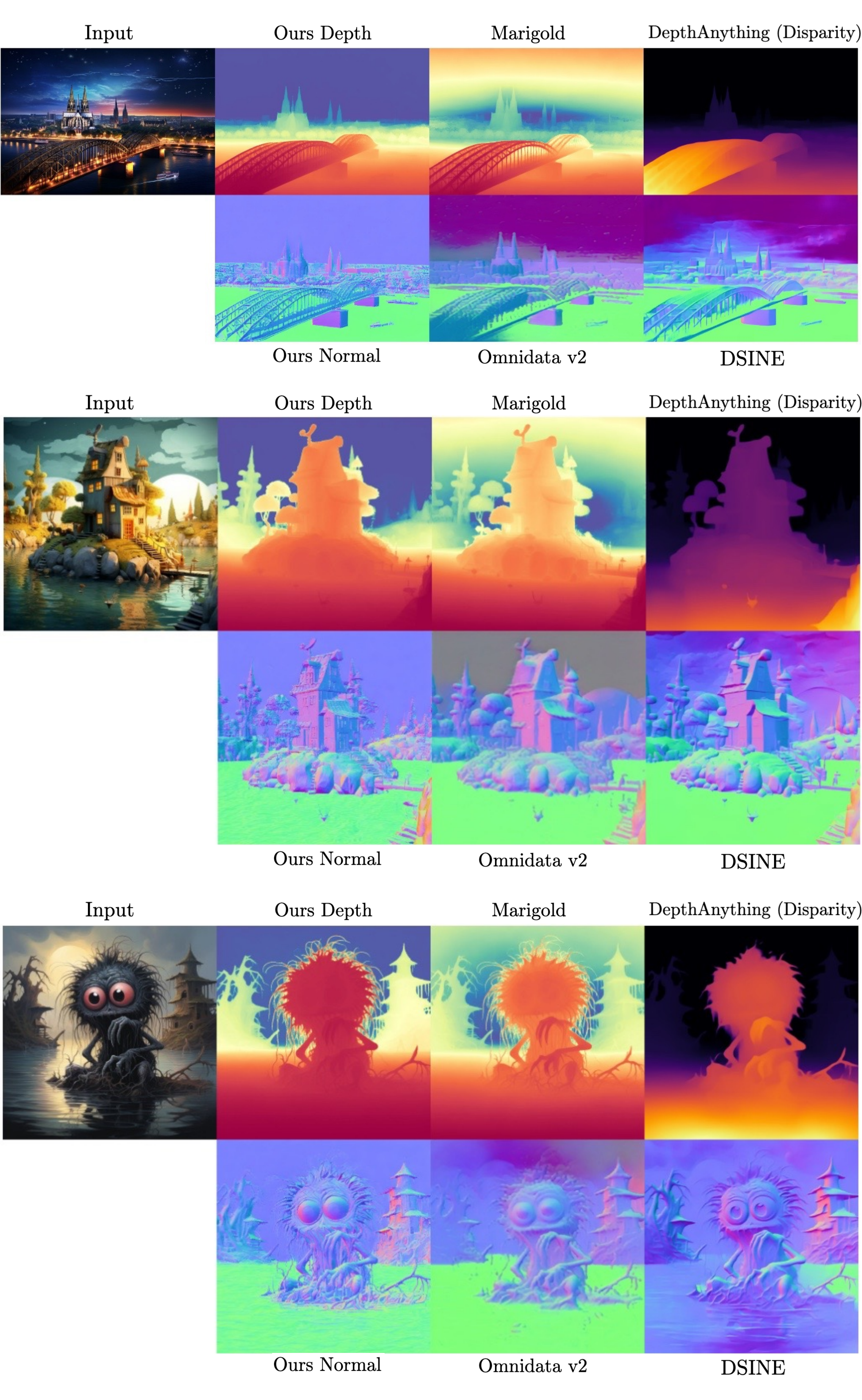}
\caption{Qualitative geometry comparison on in-the-wild images (11/11).}
\label{fig:inthewild_comp_11}
\end{figure*}

We collect in-the-wild images that are publicly available and allow for disclosure from the Internet, our daily life, or AI-generated pool to test the generalizability. For examples in the main paper, we carefully transform each inverse relative depth to relative depth with manually estimated scale and shift for clearer differentiation. To prevent any confusion regarding this transformation, we maintain the original color bar in the disparity depth maps in the supplementary, and this still demonstrates obvious differences in high-frequency details. As shown from~\figref{fig:inthewild_comp_1} to~\figref{fig:inthewild_comp_11}, GeoWizard consistently produces high-fidelity details and correct spatial layout compared to baselines,~\ie, Marigold and DepthAnything for depth, and Omnidata v2 and DSINE for normal.

\begin{figure*}[h]
\centering
\includegraphics[width=.98\linewidth]{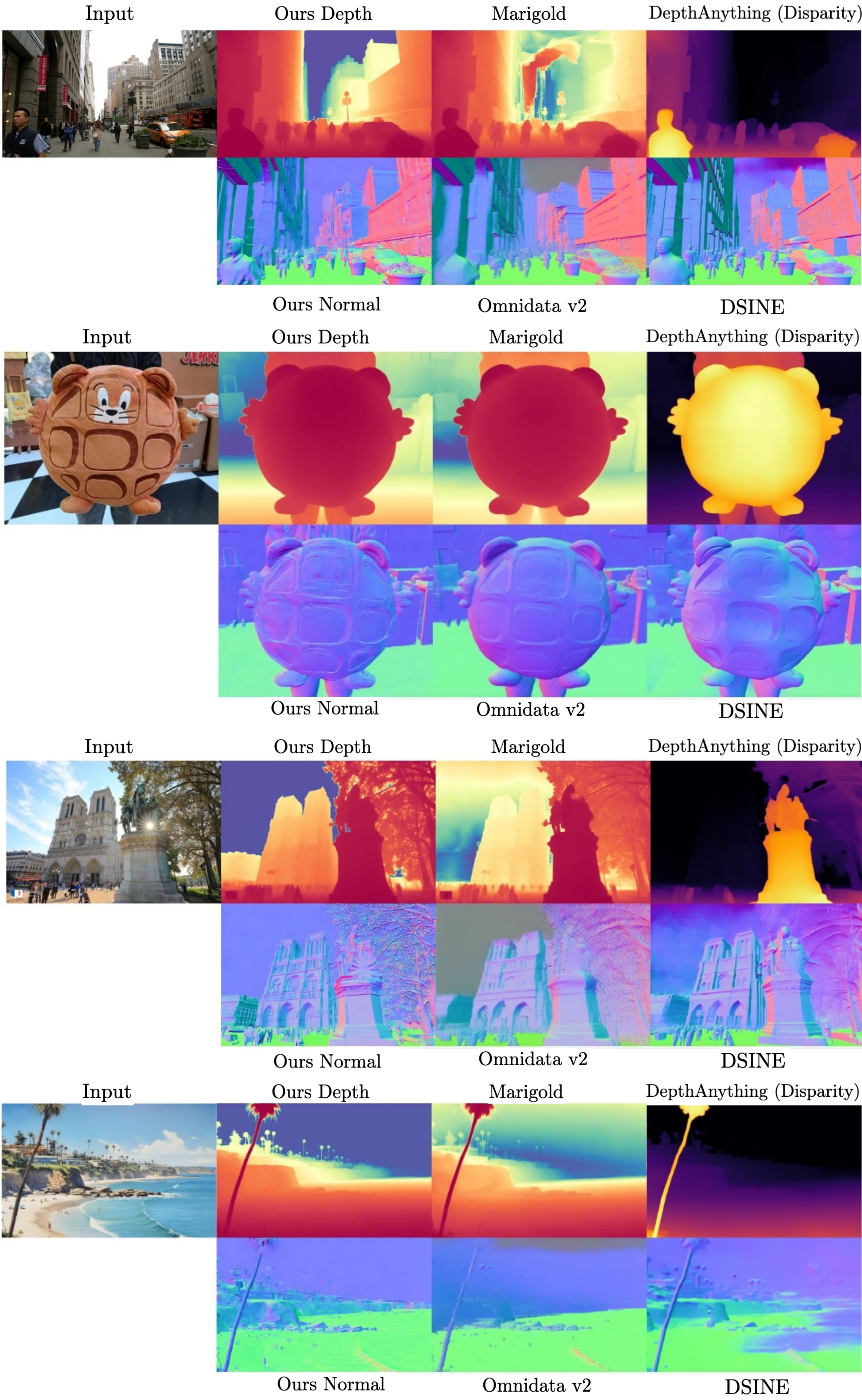}
\caption{Qualitative geometry comparison on in-the-wild images (2/11).}
\end{figure*}

\begin{figure*}[h]
\centering
\includegraphics[width=.98\linewidth]{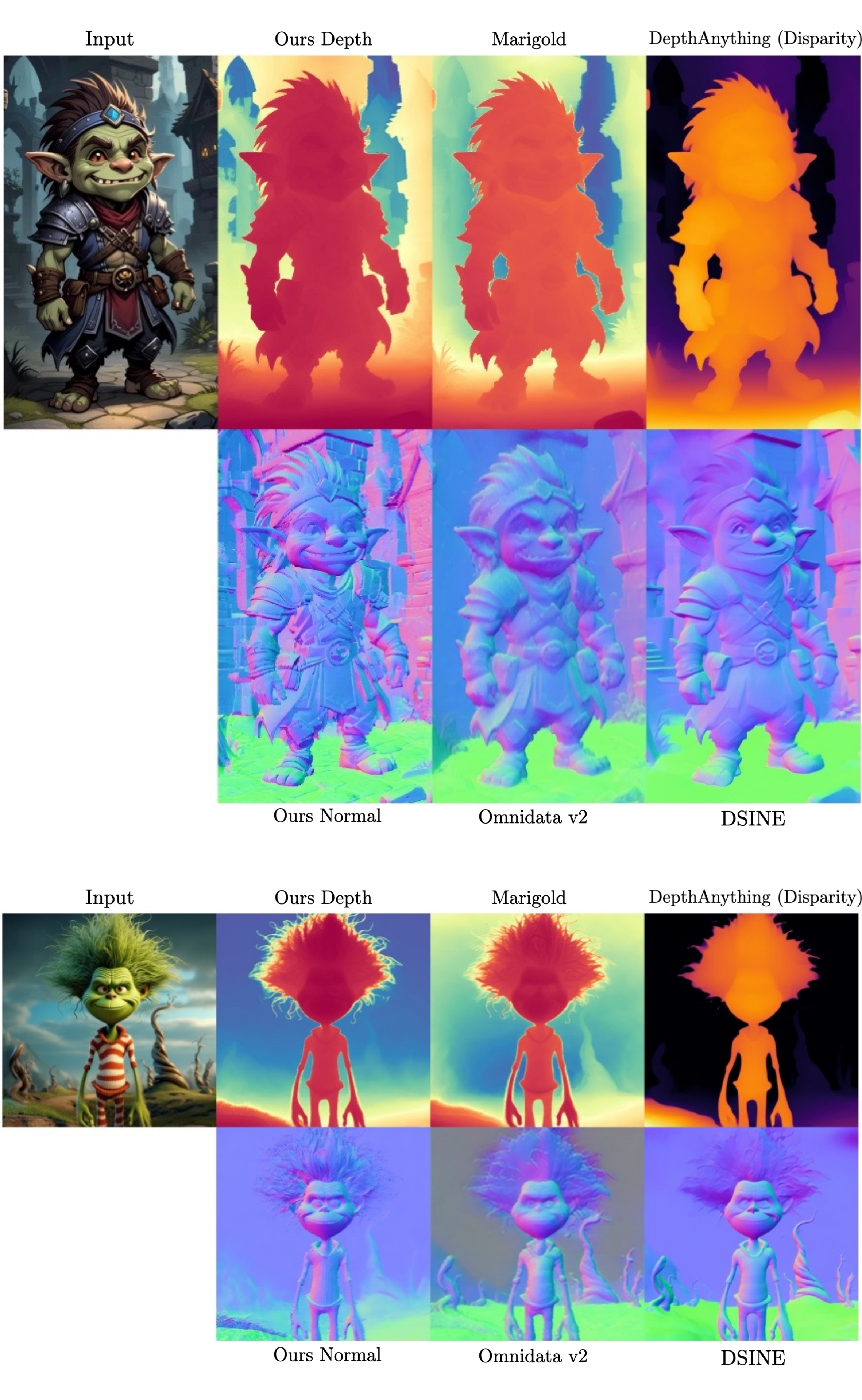}
\caption{Qualitative geometry comparison on in-the-wild images (3/11).}
\end{figure*}

\begin{figure*}[h]
\centering
\includegraphics[width=.98\linewidth]{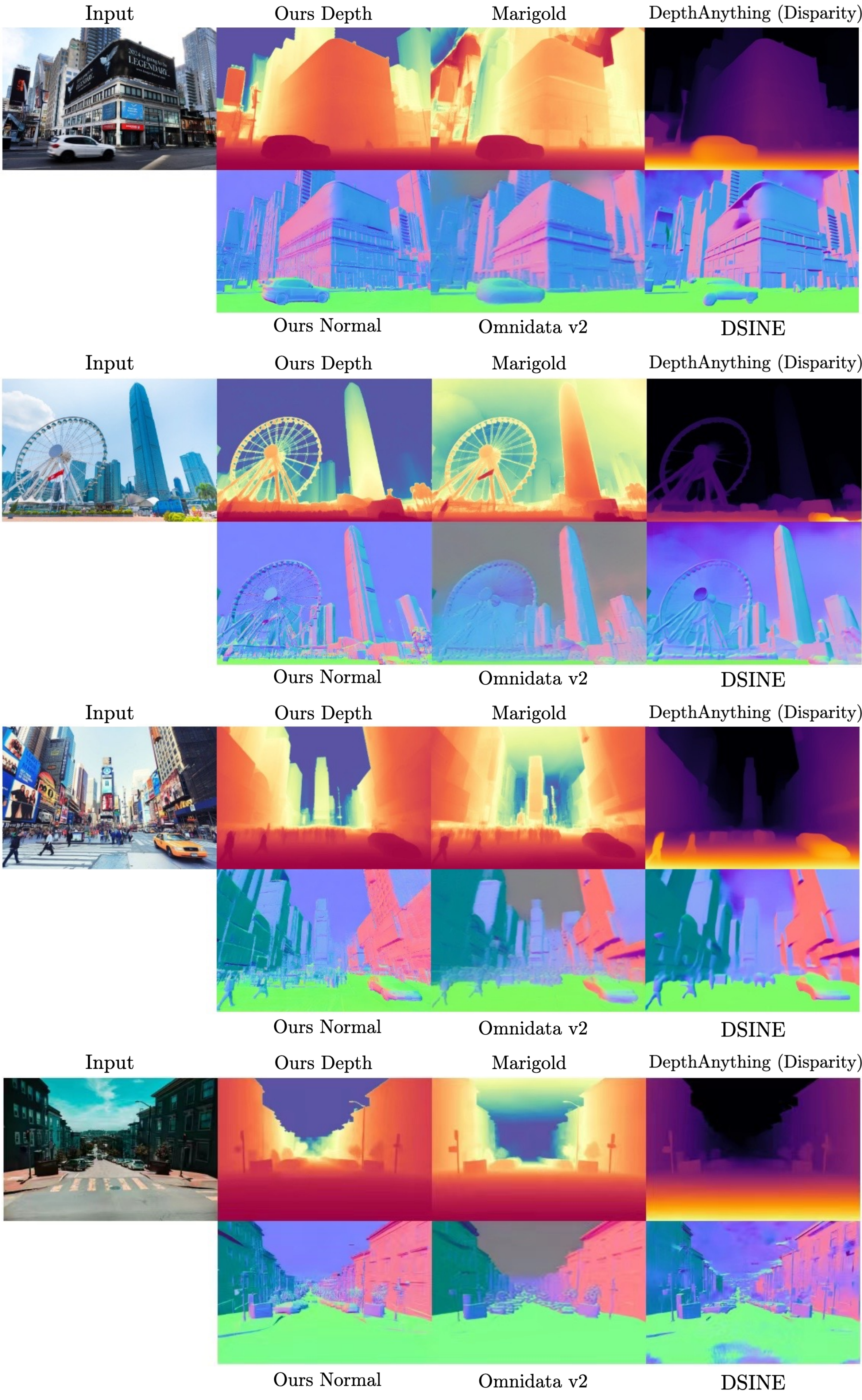}
\caption{Qualitative geometry comparison on in-the-wild images (4/11).}
\end{figure*}

\begin{figure*}[h]
\centering
\includegraphics[width=.98\linewidth]{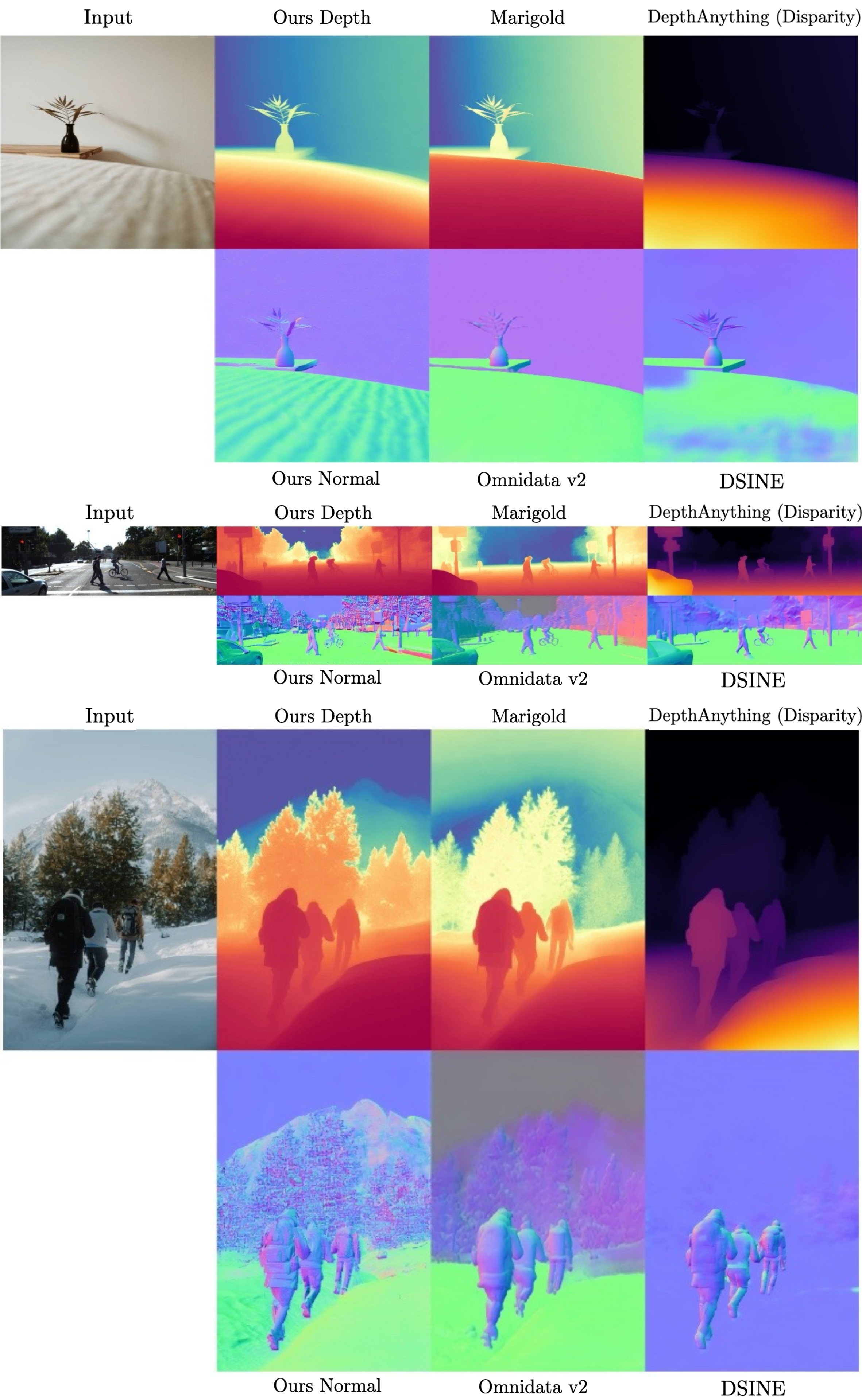}
\caption{Qualitative geometry comparison on in-the-wild images (5/11).}
\end{figure*}

\begin{figure*}[h]
\centering
\includegraphics[width=.98\linewidth]{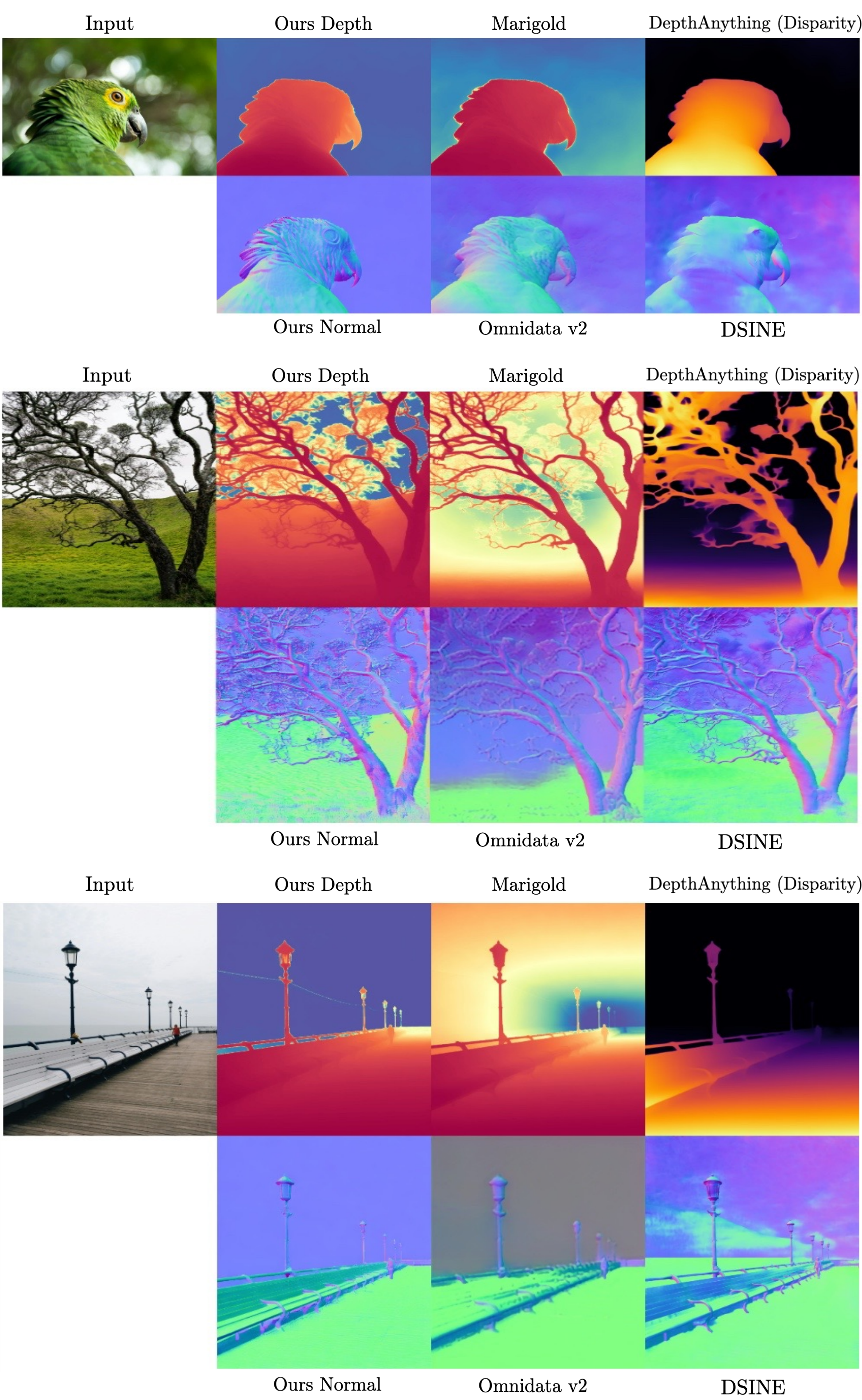}
\caption{Qualitative geometry comparison on in-the-wild images (6/11).}
\end{figure*}

\begin{figure*}[h]
\centering
\includegraphics[width=.98\linewidth]{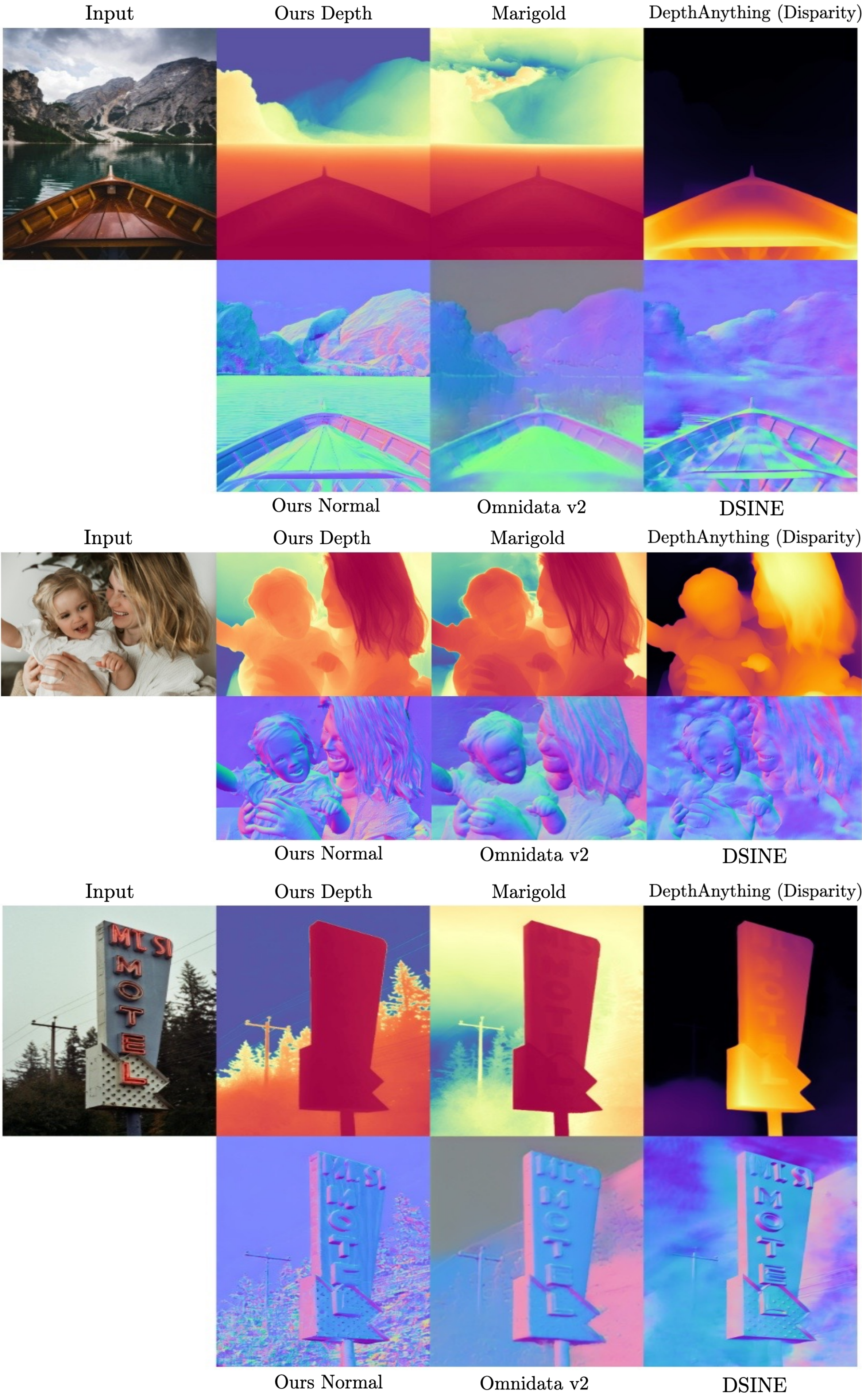}
\caption{Qualitative geometry comparison on in-the-wild images (7/11).}
\end{figure*}

\begin{figure*}[h]
\centering
\includegraphics[width=.98\linewidth]{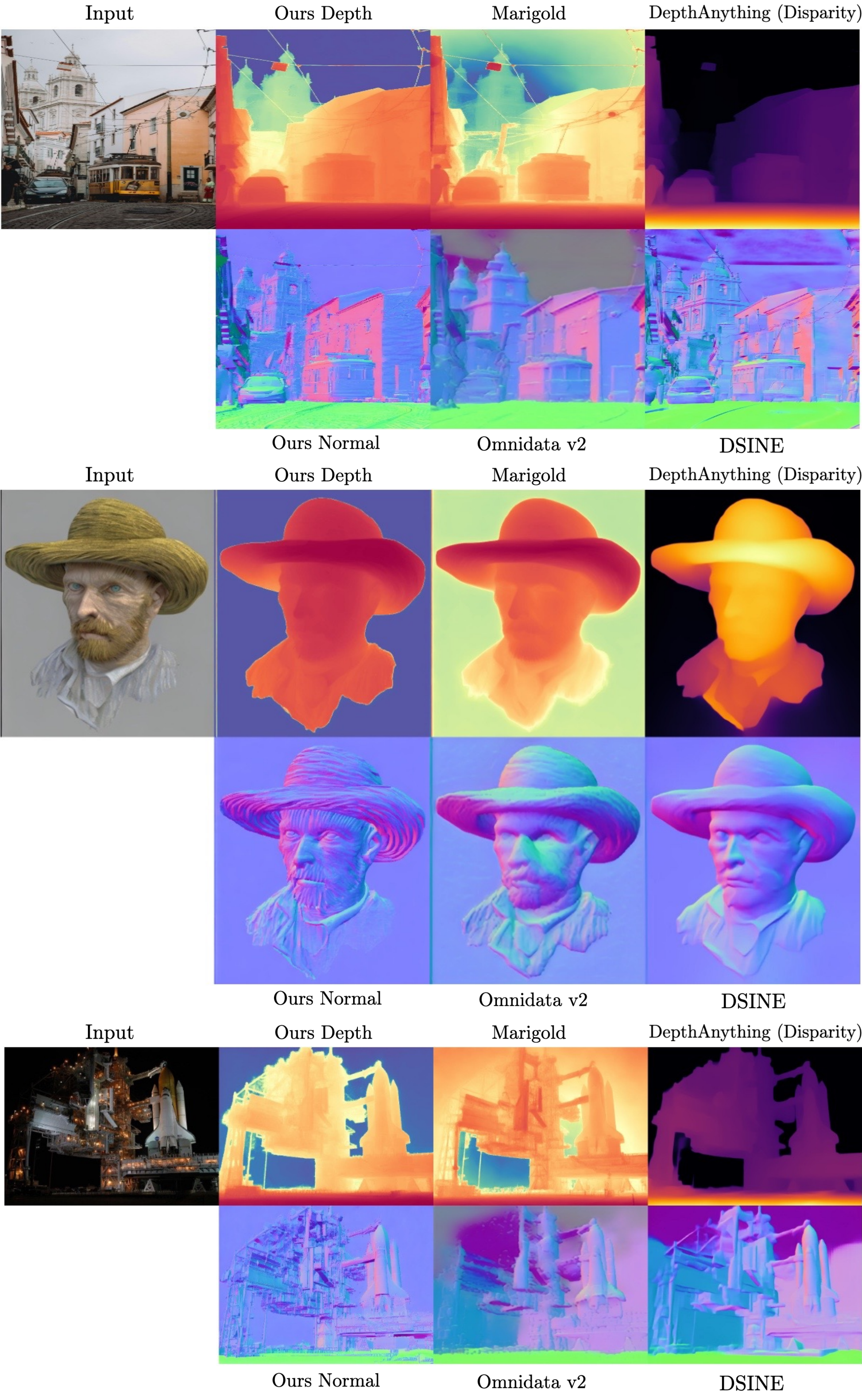}
\caption{Qualitative geometry comparison on in-the-wild images (8/11).}
\end{figure*}

\begin{figure*}[h]
\centering
\includegraphics[width=.98\linewidth]{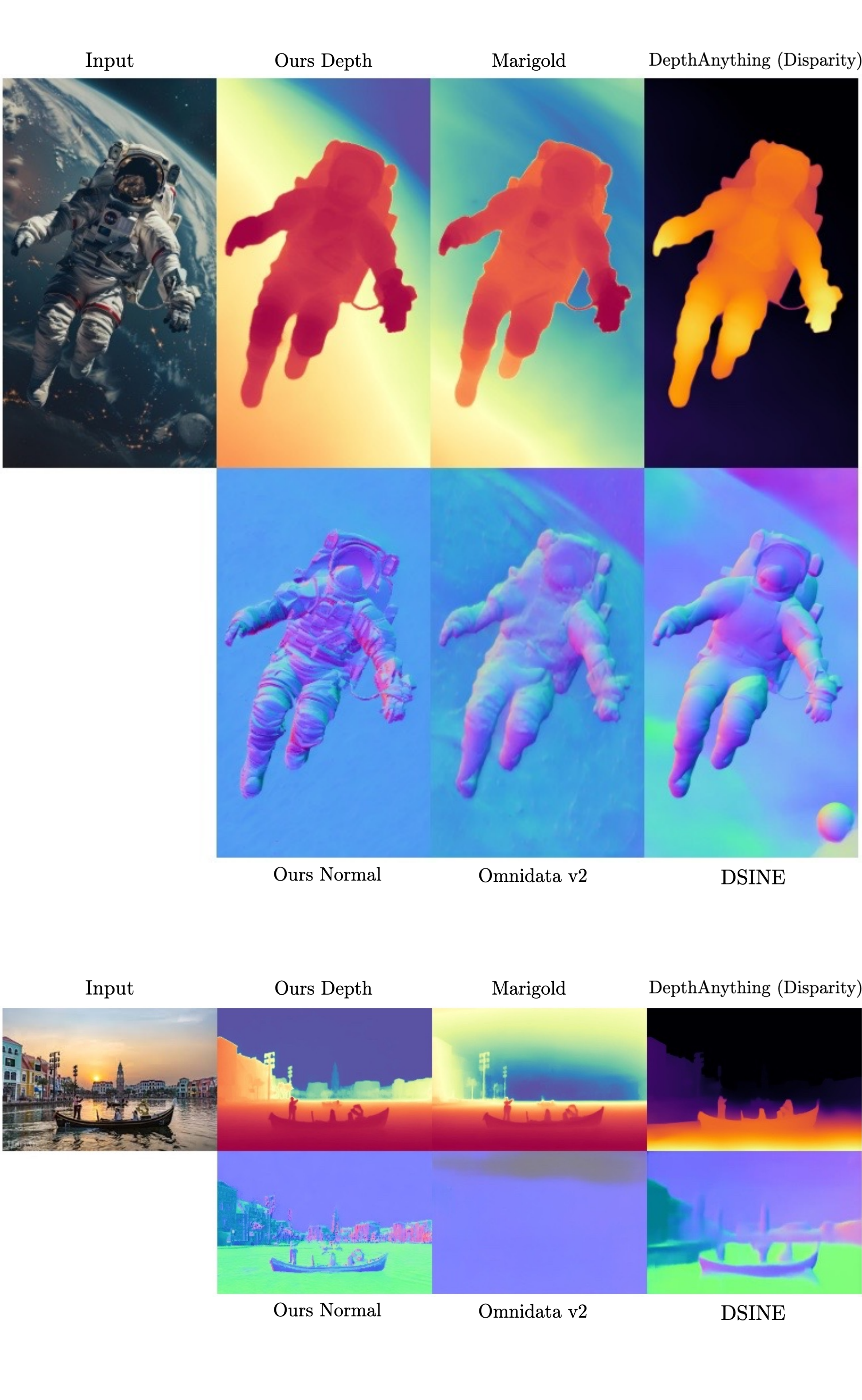}
\caption{Qualitative geometry comparison on in-the-wild images (9/11).}
\end{figure*}

\begin{figure*}[h]
\centering
\includegraphics[width=.98\linewidth]{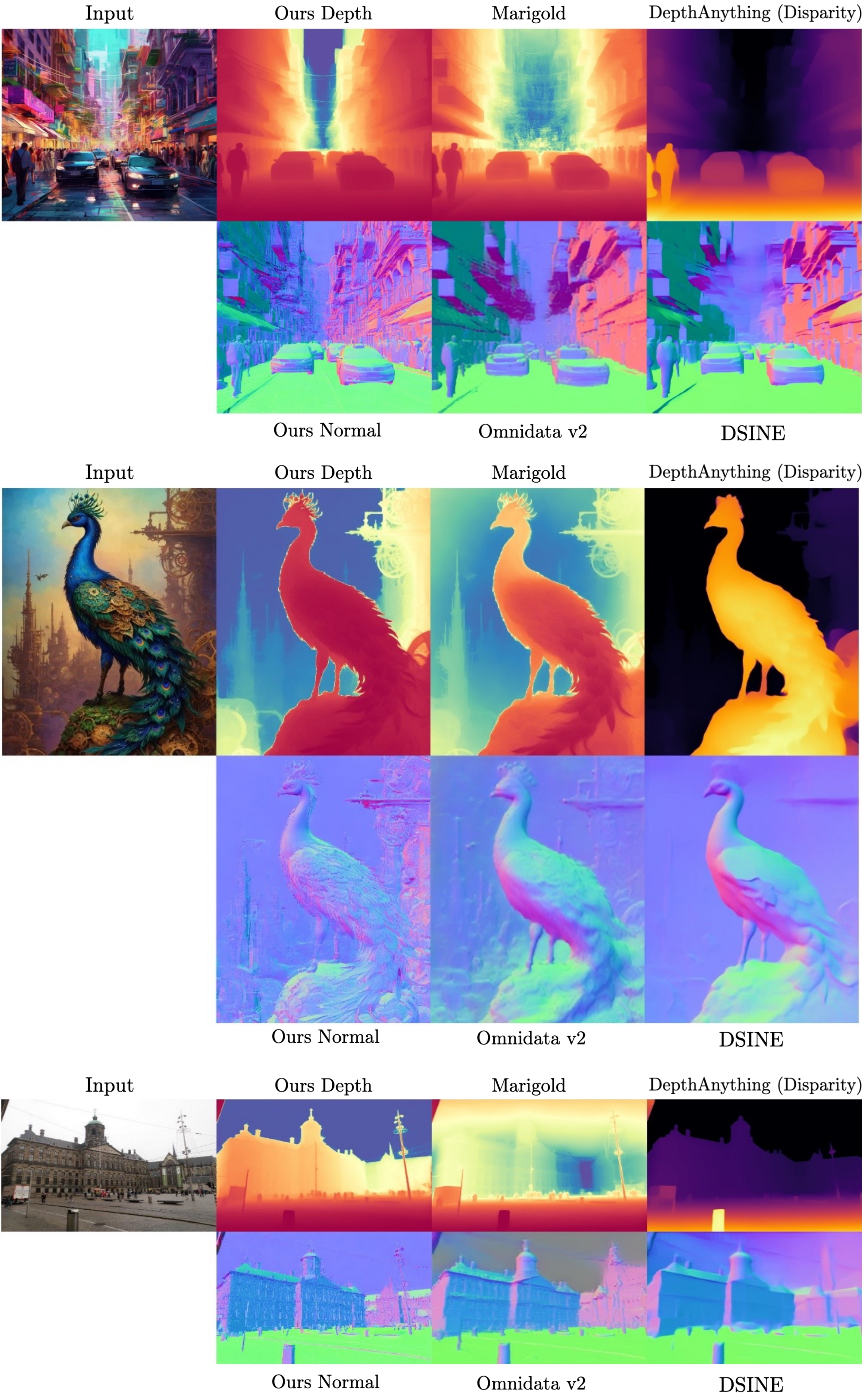}
\caption{Qualitative geometry comparison on in-the-wild images (10/11).}
\end{figure*}

\subsection{In-the-Wild 3D Reconstruction}
\begin{figure*}[h]
\centering
\includegraphics[width=\linewidth]{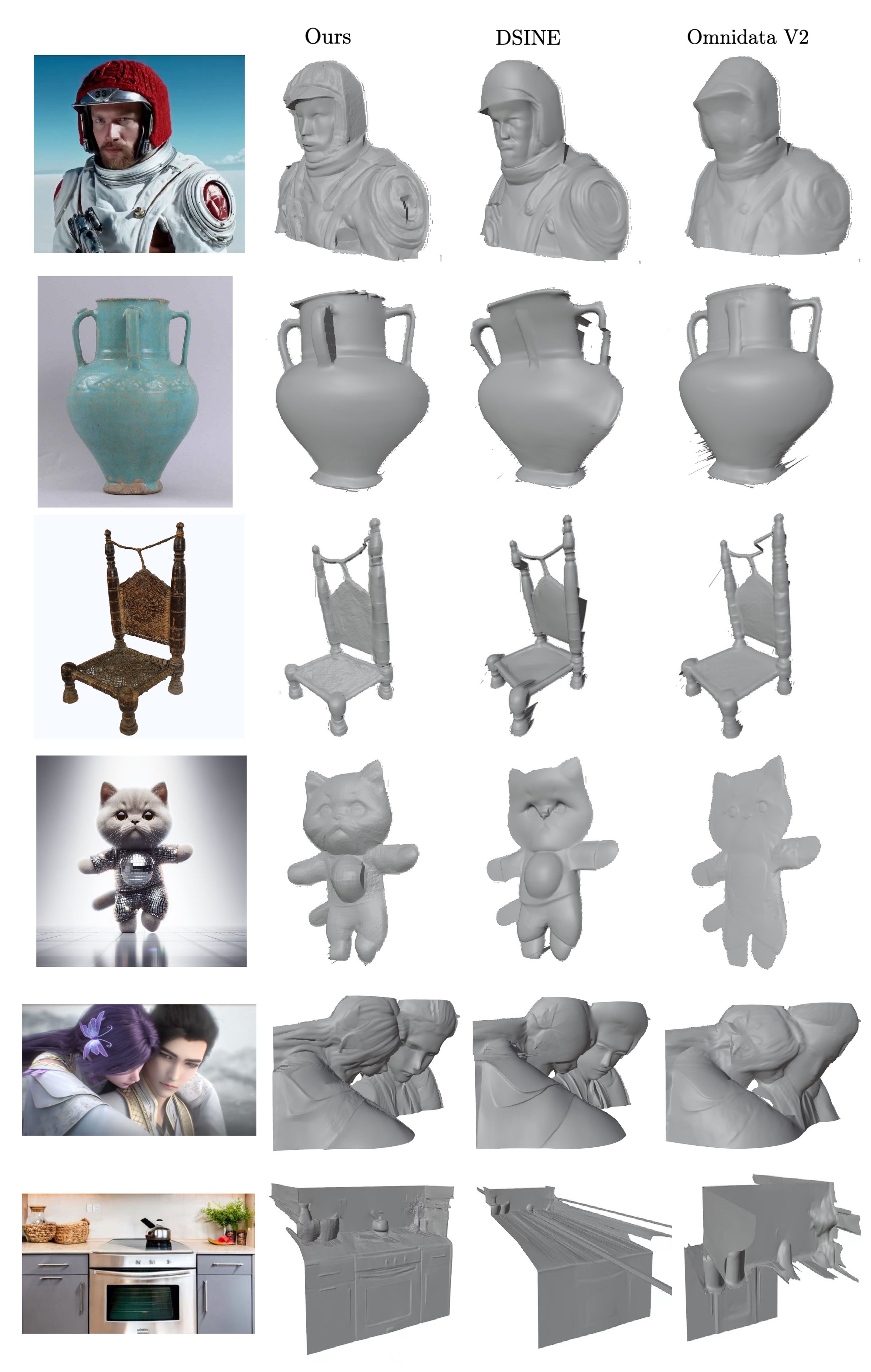}
\caption{Qualitative comparison on 3D reconstruction. We segment out the foreground objects using SAM~\cite{kirillov2023segment}. The meshes rotate left and right along the z-axis.}
\label{fig:supp_wild_recon}
\end{figure*}

We provide more 3D reconstruction results as visualized in~\figref{fig:supp_wild_recon}, comparing Ours with DSINE~\cite{bae2024dsine} and Omnidata v2~\cite{kar20223d}. For a fair comparison, we exclusively use only normal maps as input for the BiNI algorithm~\cite{cao2022bilateral}. The meshes reconstructed by GeoWizard generate enhanced high-frequency details, including hair, clothing folds, metal and wood textures, and thin handrails. Meanwhile, it delivers superior predictions of the 3D structural layout that align more closely with the original input image.

\subsection{Depth-aware Novel View Synthesis}

\begin{figure*}[h]
\centering
\includegraphics[width=\linewidth]{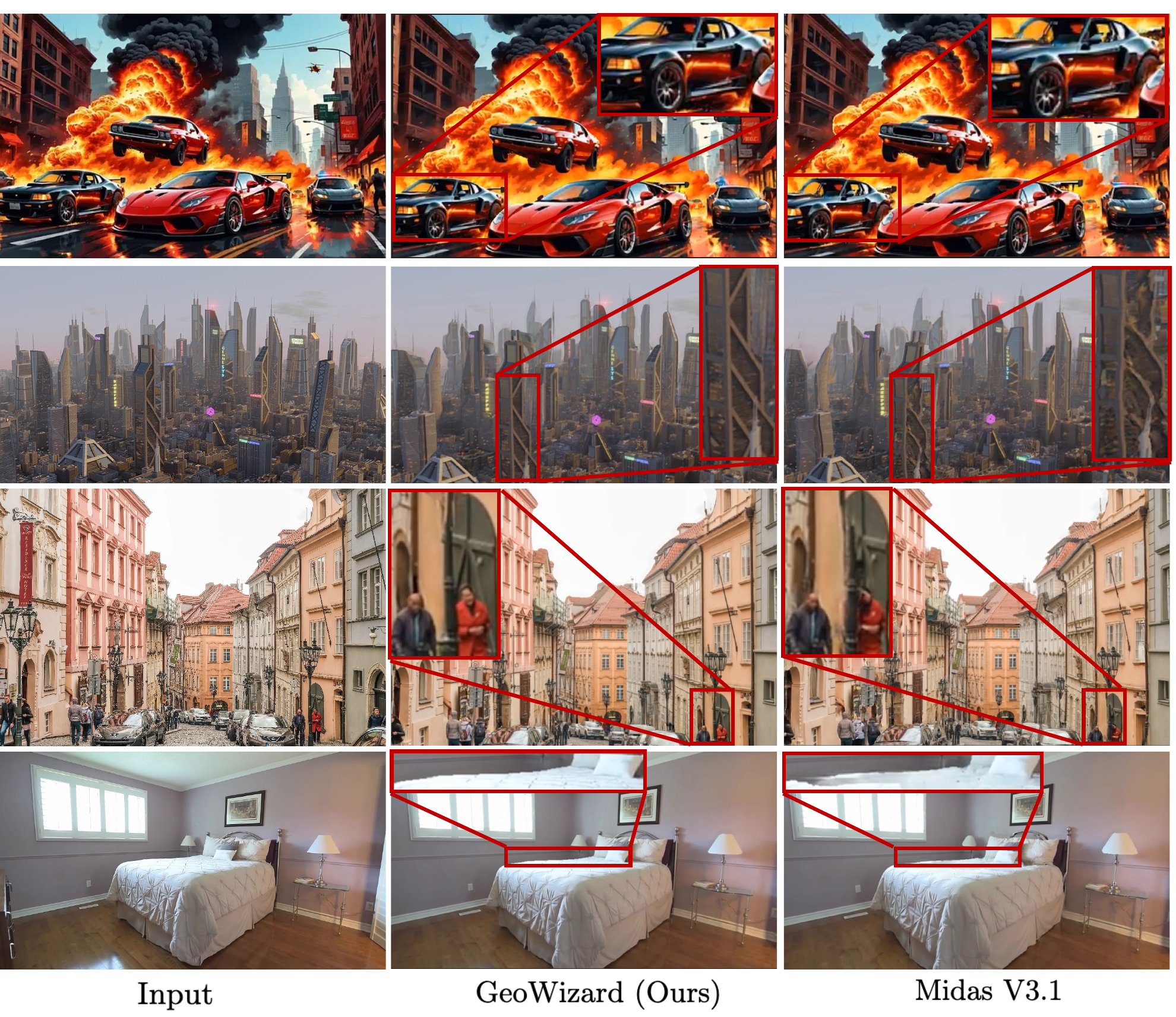}
\caption{Novel view synthesis comparison on more scenes.}  
\label{fig:supp_novel_view_synthesis}
\end{figure*}

We present more novel view synthesis results as shown in~\figref{fig:supp_novel_view_synthesis}. Our approach, GeoWizard, outperforms Midas V3.1~\cite{Ranftl2022} to guide the generation of more coherent and believable structures for objects that pose challenges in monocular depth estimation, including AI generated cars, buildings with unusual shapes, slender lampposts, and white bed under sunlight. Since this method~\cite{shih20203d} takes inverse depth in pretraining, thus the manual transformation of our depth into its inverse form will cause accuracy loss. And we find the difference in the novel views generated by our model compared to DepthAnything is relatively minor.

\section{Limitation and Potential Negative Social Impact}  \label{supp:limitation}
GeoWizard serves as a foundation model for estimating geometry in both real-world and artifically created images. Despite its strengths, the current framework still has some limitations. First, the iterative denoising process is time-consuming when applied to large-scale collections. Since the depth and normal maps are generated from randomly initialized noise, this diffusion leads to inconsistencies when applied to video sequences. In terms of the reconstruction, the pseudo scale and shift derived from the combined depth and normal maps may exhibit accuracy issues in some cases. Meanwhile, some concerns exist when making our models publicly available. It model can be extended to create fake but realistic 3D assets. Depth and normal maps play important roles in scene understanding, and our model could be incorporated into surveillance systems to identify regions that are not clearly distinguishable to the human eyes. To mitigate these issues, we will include stipulations in the license agreement for the code limiting its applications only to academic research.

%% file: tabs/ablation_supp.tex
\begin{table}[h]
\centering
\resizebox{\textwidth}{!} 
{
\begin{tabular}{c|ccc|ccc|ccc|ccc}
\toprule
\multirow{2}{*}{Method} & \multicolumn{3}{c}{ Indoor } & \multicolumn{3}{c}{Outdoor} & \multicolumn{3}{c}{Object} & \multicolumn{3}{c}{Overall} \\

& AbsRel $\downarrow$ & Mean $\downarrow$ & GC $\downarrow$ & AbsRel $\downarrow$ & Mean $\downarrow$ & GC $\downarrow$ & AbsRel $\downarrow$ & Mean $\downarrow$ & GC $\downarrow$ & AbsRel $\downarrow$ & Mean $\downarrow$ & GC $\downarrow$ \\

\midrule

w/ Indoor Indicator & \textbf{5.5} & \textbf{12.6} & 14.7 & 10.1 & 22.8 & 23.9 & 3.7 & 15.8 & 17.7 & 6.8 & 15.0 & 16.4 \\
w/ Outdoor Indicator & 5.8 & 13.1 & \textbf{14.4} & \textbf{9.6} & \textbf{22.1} & \textbf{23.5} & 3.9 & 15.9 & 18.2 & 7.0 & 15.2 & 16.4 \\
w/ Object Indicator & 6.4 & 13.7 & 14.9 & 10.8 & 23.5 & 23.7 & 3.5  & \textbf{15.4} & \textbf{17.6} & 7.5 & 15.5 & 16.6 \\

Shared Geometry~\cite{liu2023hyperhuman} & 6.1 & 13.2 & 14.6 & 10.4 & 23.6 & 23.8 & 3.6 & 16.4 & 17.8 & 7.2 & 15.3 & 16.3 \\
\midrule

Full Model & \textbf{5.5} & \textbf{12.6} & 14.7 & \textbf{9.6} & \textbf{22.1} & \textbf{23.5} & 3.5  & \textbf{15.4} & \textbf{17.6} & \textbf{6.7} & \textbf{14.8} & \textbf{16.2} \\

\bottomrule
\end{tabular}
}

\caption{Quantitative ablation studies on different scene types.} 
\label{tab:ablation_supp}
\vspace{-2em}
\end{table}